\documentclass[letterpaper]{article} 
\usepackage{aaai2026}  
\usepackage{times}  
\usepackage{helvet}  
\usepackage{courier}  
\usepackage[hyphens]{url}  
\usepackage{graphicx} 
\usepackage{natbib}  
\usepackage{caption} 
\usepackage{algorithm}
\usepackage{algorithmic}
\usepackage{amsmath}
\usepackage{amssymb}

\usepackage{newfloat}
\usepackage{listings}
\DeclareCaptionStyle{ruled}{labelfont=normalfont,labelsep=colon,strut=off} 
\floatstyle{ruled}
\newfloat{listing}{tb}{lst}{}
\floatname{listing}{Listing}
\title{Hyperspectral Variational Autoencoders\\for Joint Data Compression and Component Extraction}
\author {
    Core Francisco Park\textsuperscript{\rm 1,\rm 2,\rm 3,\rm 4},
    Manuel Pérez-Carrasco\textsuperscript{\rm 3},
    Caroline Nowlan\textsuperscript{\rm 3},
    Cecilia Garraffo\textsuperscript{\rm 3,\rm 4}
}
\affiliations {
    \textsuperscript{\rm 1}CBS-NTT Program in Physics of Intelligence, Harvard University, Cambridge, MA, USA\\
    \textsuperscript{\rm 2}Center for Brain Science, Harvard University, Cambridge, MA, USA\\
    \textsuperscript{\rm 3}Center for Astrophysics $|$ Harvard \& Smithsonian, Cambridge, MA, USA\\
    corefranciscopark@g.harvard.edu, cgarraffo@cfa.harvard.edu
}

\usepackage{bibentry}

\begin{document}

\maketitle

\begin{abstract}
Geostationary hyperspectral satellites generate terabytes of data daily, creating critical challenges for storage, transmission, and distribution to the scientific community. We present a variational autoencoder (VAE) approach that achieves $\times$514 compression of NASA's TEMPO satellite hyperspectral observations (1028 channels, 290-490nm) with reconstruction errors 1-2 orders of magnitude below the signal across all wavelengths. This dramatic data volume reduction enables efficient archival and sharing of satellite observations while preserving spectral fidelity. Beyond compression, we investigate to what extent atmospheric information is retained in the compressed latent space by training linear and nonlinear probes to extract Level-2 products (NO$_2$, O$_3$, HCHO, cloud fraction). Cloud fraction and total ozone achieve strong extraction performance (R$^2$=0.93 and 0.81 respectively), though these represent relatively straightforward retrievals given their distinct spectral signatures. In contrast, tropospheric trace gases pose genuine challenges for extraction (NO$_2$ R$^2$=0.20, HCHO R$^2$=0.51) reflecting their weaker signals and complex atmospheric interactions. Critically, we find the VAE encodes atmospheric information in a semi-linear manner—nonlinear probes substantially outperform linear ones—and that explicit latent supervision during training provides minimal improvement, revealing fundamental encoding challenges for certain products. This work demonstrates that neural compression can dramatically reduce hyperspectral data volumes while preserving key atmospheric signals, addressing a critical bottleneck for next-generation Earth observation systems.
\end{abstract}

\begin{links}
     \link{Code}{https://github.com/cfpark00/Hyperspectral-VAE}
\end{links}

\section{Introduction}

\begin{figure*}[t!]
    \centering
    \includegraphics[width=\textwidth]{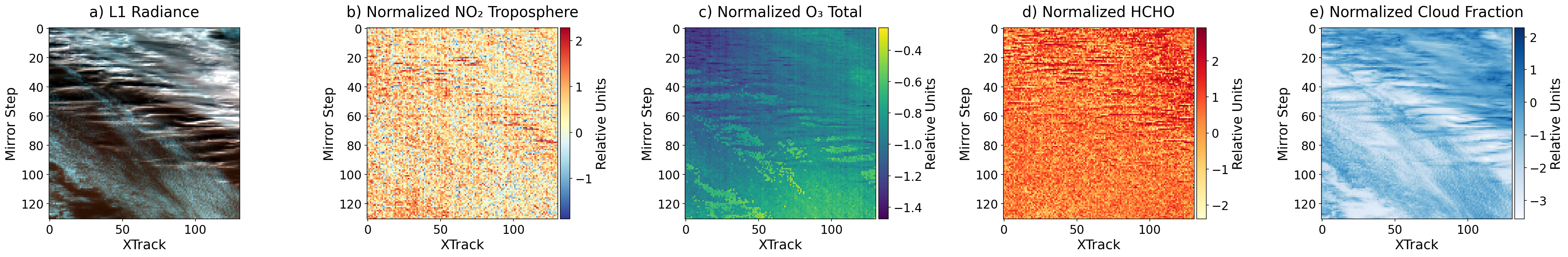}
    \caption{\textbf{TEMPO satellite data products.} Representative $131 \times 131$ pixel region showing (a) Level-1 radiance data as three-channel composite (channels 100, 500, 900 out of 1028), and (b-e) Level-2 atmospheric products: tropospheric NO$_2$, total ozone, HCHO, and cloud fraction. All L2 products are normalized (see App.~\ref{app:data_norm}). The distinct spatial patterns—from localized NO$_2$ pollution to broad stratospheric ozone—illustrate the challenge of preserving multiple atmospheric signals through compression.}
    \label{fig:tempo_data}
\end{figure*}

Atmospheric monitoring from space has become essential for understanding air quality, tracking pollution sources, and studying climate change \citep{goetz1985imaging,thies2011satellite,holloway2021satellite,yang2013role,weng2009thermal}.
The recently launched Tropospheric Emissions: Monitoring of Pollution (TEMPO) satellite \citep{zoogman2017tropospheric} represents a major advance in this capability, providing hourly observations of atmospheric composition over North America with unprecedented spatial and spectral resolution.
TEMPO's hyperspectral measurements span 1028 channels across each of the UV-visible (293 - 494 nm) and visible-near-infrared spectrum (538 - 741 nm), enabling detailed characterization of trace gases like nitrogen dioxide (NO$_2$), ozone (O$_3$), and formaldehyde (HCHO).
However, hyperspectral satellite observations generate massive data volumes that pose significant challenges for storage, transmission, and sharing.
Unlike conventional imaging where each pixel contains a small number of channels (most often RGB), hyperspectral data encodes a full spectrum at every spatial location, resulting in orders of magnitude larger file sizes.
TEMPO's geostationary orbit compounds this issue by enabling continuous hourly observations, generating terabytes of data daily coming from 2.4 million spectra every hour that must be archived, distributed to researchers, and processed for operational products.

Traditional physics-based retrieval algorithms \citep{rodgers2000inverse,bioucas2013hyperspectral} address dimensionality through carefully selected spectral windows and prior constraints, but these approaches require extensive domain expertise and remain computationally intensive \citep{zoogman2017tropospheric,chance2013tropospheric}.
Deep learning offers an alternative paradigm that can learn efficient compressed representations directly from raw spectral measurements.
Variational Autoencoders (VAEs) \citep{kingma2013auto,higgins2016beta} are particularly well-suited for this task as they provide unsupervised dimensionality reduction while learning interpretable latent representations that can be probed for physical quantities.

We present a VAE-based approach that achieves $\times$514 compression of TEMPO hyperspectral data with minimal spectral reconstruction error, addressing the critical challenge of storing and sharing massive satellite observation volumes.
Our method jointly encodes spatial and spectral dimensions, compressing raw $[1028 \times 64 \times 64]$ radiance tiles into compact $[32 \times 16 \times 16]$ latent representations while accurately preserving spectral features across the full 290-490nm range.
We then employ a two-stage validation strategy: first training the VAE for reconstruction, then using linear and nonlinear probes to extract Level-2 atmospheric products (NO$_2$, O$_3$, HCHO, and cloud fraction) from the latent space.
We find that the VAE naturally encodes atmospheric information in a semi-linear manner—linear probes extract some information, but nonlinear MLP probes substantially outperform them, revealing that atmospheric products are not directly accessible through simple linear combinations of latent dimensions.
Surprisingly, explicitly supervising the VAE latent space with L2 product predictions during training provides minimal improvement over unsupervised compression, suggesting fundamental challenges in directly encoding certain atmospheric components regardless of supervision strategy.

The compression achieves excellent spectral fidelity, with reconstruction errors 1-2 orders of magnitude smaller than the signal across all 1028 channels.
Beyond preserving spectral structure, the learned latent representations encode atmospheric information with varying degrees of success across different products.
Cloud fraction and total ozone achieve strong extraction performance (R$^2$=0.92 and 0.75 respectively) with MLP probes, indicating that the VAE effectively captures cloud patterns and stratospheric composition.
Tropospheric trace gases NO$_2$ and HCHO prove more challenging (R$^2$=0.21 and 0.50), though the substantial improvements from linear (R$^2$=0.12 and 0.47) to nonlinear probing demonstrate that information exists but requires nonlinear extraction.
Critically, training a supervised VAE variant that explicitly predicts L2 products from the latent space yields nearly identical probe performance, revealing that the difficulty lies in the fundamental encoding challenge rather than lack of supervision.
These results demonstrate that unsupervised compression can preserve key atmospheric signals in hyperspectral data, with implications for efficient storage and transmission of satellite observations.

Our main contributions are:
\begin{itemize}
\item We achieve $\times$514 compression of TEMPO hyperspectral satellite data with minimal spectral reconstruction error (1-2 orders of magnitude below signal), demonstrating that deep learning can dramatically reduce storage and transmission requirements while preserving spectral fidelity.
\item We demonstrate that unsupervised learning can encode certain atmospheric products in the latent space, with cloud fraction and total ozone achieving R$^2$ scores of 0.92 and 0.75 respectively, while tropospheric trace gases remain challenging.
\item We provide a comprehensive analysis comparing linear and nonlinear extraction methods, revealing that the VAE encodes atmospheric information in a semi-linear manner where nonlinear probes substantially outperform linear ones, and that explicit latent supervision provides minimal benefit.
\item We establish a framework for joint spatial-spectral compression of satellite observations that maintains both spectral fidelity and spatial coherence, and introduce a latent-supervised architecture that enables joint optimization of compression and atmospheric product extraction.
\end{itemize}

The remainder of this paper is organized as follows.
Section 2 reviews related work in hyperspectral image analysis and machine learning for atmospheric remote sensing.
Section 3 describes our methodology, including the VAE architecture, data preprocessing pipeline, and probe analysis framework.
Section 4 presents experimental results comparing reconstruction quality and component extraction performance across different atmospheric products.
Section 5 concludes with a discussion of implications for operational satellite data processing and directions for future research.

\begin{figure*}[ht]
    \centering
    \includegraphics[width=\textwidth]{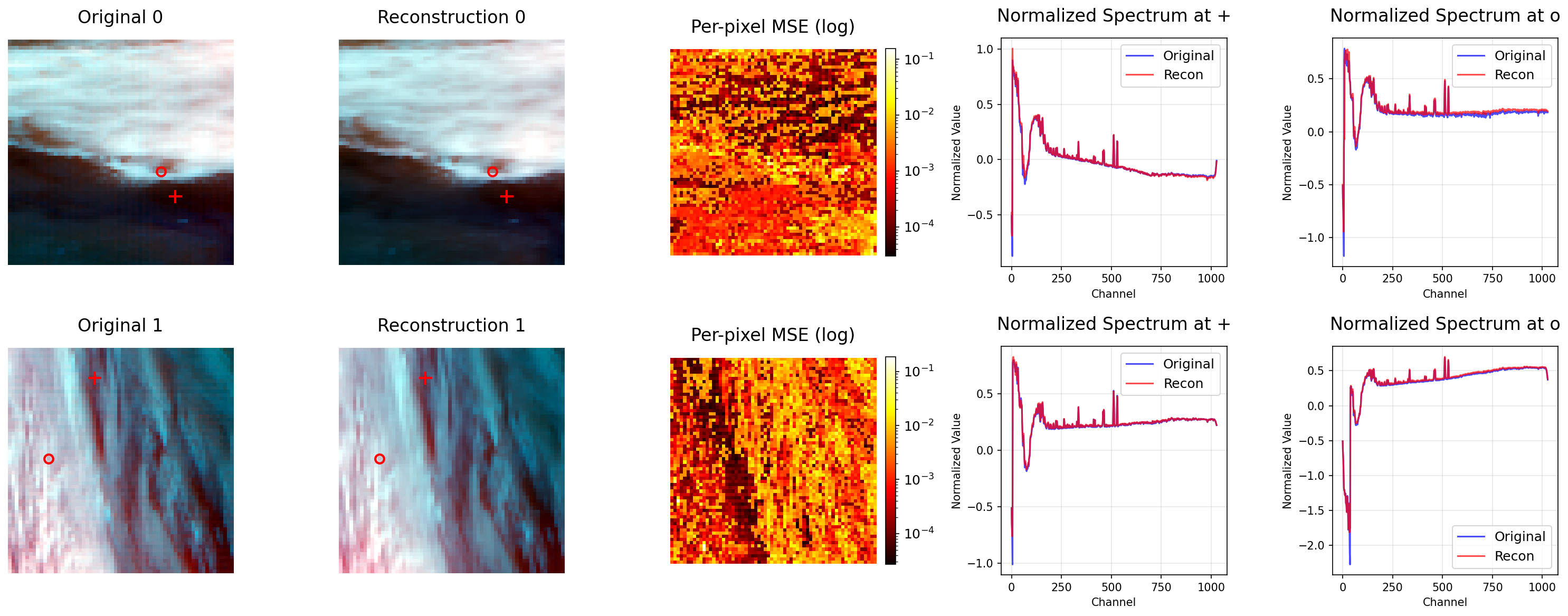}
    \caption{\textbf{VAE reconstruction quality.} Two representative samples showing VAE reconstruction performance at 200,000 training steps. For each sample, we display: (left) original TEMPO radiance as three-channel composite (channels 100, 500, 900), (center-left) VAE reconstruction, (center) per-pixel MSE on log scale, and (right panels) normalized spectra at two random spatial locations (marked with red + and o). The VAE achieves $\times$514 compression ($1028 \times 64 \times 64 \rightarrow 32 \times 16 \times 16$) while accurately preserving both spatial structures and spectral features across the 290-490nm range. Note: The large features visible in low-index channels (left end, near 290 nm) are primarily artifacts from measurement uncertainties and stray light rather than genuine atmospheric signals.}
    \label{fig:reconstruction}
    \end{figure*}

\section{Related Work}

\subsection{Neural Image Compression}
Recent advances in learned image compression have demonstrated that neural networks can achieve superior rate-distortion performance compared to traditional codecs. Variational autoencoders and generative models have emerged as powerful frameworks for lossy compression, learning compact representations that preserve perceptually important features while achieving high compression ratios \citep{yang2024lossyimagecompressionconditional}. \citet{Li2024.07.07.601368} demonstrated the potential of neural compression for volumetric biological data, achieving significant compression while maintaining semantic information. Hyperspectral satellite data presents an ideal application domain for neural compression due to strong correlations across both spectral and spatial dimensions—neighboring wavelengths exhibit smooth spectral features while adjacent pixels share similar atmospheric conditions, creating natural redundancy that can be exploited for compression. Unlike natural images where perceptual quality drives optimization, satellite data compression must maintain quantitative accuracy for scientific retrievals, requiring careful consideration of spectral fidelity and spatial coherence across the full wavelength range.

\subsection{VAEs and Disentangled Representations}
Variational autoencoders have proven effective at learning disentangled representations \citep{kingma2013auto,higgins2016beta} that separate underlying factors of variation in complex data. The VAE framework's probabilistic nature provides a principled approach to representation learning, with the KL divergence regularization encouraging compact, structured latent spaces. In the context of atmospheric observations, disentanglement could be particularly valuable as different atmospheric constituents have distinct spectral signatures that might be encoded in separate latent dimensions. Our approach leverages this capability to learn representations that facilitate extraction of specific atmospheric products, though we find that purely unsupervised learning has limitations for trace gas signals with low signal-to-noise ratios.

\subsection{Deep Learning in Remote Sensing}
The application of neural networks to satellite remote sensing has accelerated rapidly, achieving state-of-the-art performance in tasks ranging from land cover classification to atmospheric retrieval \citep{lary2016machine,mas2008application} . Convolutional architectures \cite{fukushima1980neocognitron,lecun1998convolutional,krizhevsky2009learning} have proven particularly effective for exploiting spatial correlations in satellite imagery, while attention mechanisms help capture long-range dependencies. For hyperspectral data specifically, neural approaches have shown promise in classification \citep{paoletti2019deep,manifold2021versatile}, dimensionality reduction, unmixing, target detection, and shadow removal \citep{park2023hyperspectralshadowremovaliterative,perez2025deep}. However, most existing work focuses on either spatial or spectral processing separately, whereas atmospheric remote sensing requires joint consideration of both domains. Our work addresses this gap by developing a unified spatial-spectral compression framework tailored to the unique characteristics of atmospheric satellite observations, where spectral absorption features must be preserved while leveraging spatial coherence for efficient encoding.

\section{Methodology}

\subsection{Data Preparation}

TEMPO (Tropospheric Emissions: Monitoring of Pollution) \citep{zoogman2017tropospheric} is NASA's first geostationary satellite mission dedicated to air quality monitoring over North America, providing hourly observations of atmospheric composition with unprecedented temporal resolution. The instrument measures solar radiation scattered by the atmosphere across 1028 channels in each of two spectral ranges: UV-visible (293-494 nm) and visible-near-infrared (538-741 nm), capturing key absorption features of atmospheric trace gases including NO$_2$, O$_3$, HCHO, and aerosols.

Figure \ref{fig:tempo_data} shows TEMPO Level-1 radiance data as a three-channel composite using representative spectral channels (left panel), along with Level-2 atmospheric products: tropospheric NO$_2$, total ozone, HCHO, and cloud fraction (all normalized, see Appendix~\ref{app:data_norm}). Each product exhibits distinct spatial patterns, from localized NO$_2$ pollution to broad total ozone distributions.

Our dataset consists of TEMPO Level-1B radiance measurements \citep{park2024algorithm,chong2025algorithm,nowlan2025tempo,gonzalezabad2025tempo} from January 2025 over the Los Angeles basin, a region with complex atmospheric chemistry and strong pollution gradients. The raw radiance data undergoes a multi-step normalization pipeline designed to handle the wide dynamic range typical of hyperspectral measurements: log transformation $\log(\max(\mathrm{radiance}, 1.0))$, followed by per-channel z-score normalization using global statistics computed from 2.4 million pixels across 10 representative files, and finally clipping to $[-10, 10]$ to remove outliers (see Appendix~\ref{app:data_norm} for details). We then extract $64\times64$ pixel tiles from TEMPO observations and split the data 70/30 for training and validation.

We evaluate four atmospheric products from TEMPO Level-2 data \citep{wang2025algorithm}: nitrogen dioxide (NO$_2$), total ozone (O$_3$), formaldehyde (HCHO), and cloud fraction.
Each product requires specific preprocessing to handle its physical characteristics and data distribution.
NO$_2$ and HCHO use inverse hyperbolic sine (asinh) normalization to handle negative values and heavy-tailed distributions common in trace gas measurements.
Total ozone uses standard z-score normalization as its distribution is approximately Gaussian.
Cloud fraction, bounded in $[0,1]$, uses a logit transform with squeeze: $\mathrm{logit}(0.01 + 0.98 \times \mathrm{cloud})$ to spread the values across the real line.
To align Level-2 products with the VAE's latent resolution, we apply $4\times4$ spatial pooling, reducing the $64\times64$ input tiles to $16\times16$ to match the latent spatial dimensions.

\subsection{VAE Architecture}

Our VAE architecture jointly compresses spatial and spectral dimensions through a hierarchical encoder-decoder structure with 27.3 million parameters.
The encoder progressively reduces the input tensor from shape $[1028 \times 64 \times 64]$ to a latent representation of $[32 \times 16 \times 16]$, achieving $\times$514 compression.
The encoder consists of three downsampling levels with channel dimensions $[512, 256, 128]$, where each level contains a ResNet block \citep{he2015deep} followed by $2 \times 2$ strided convolution for spatial downsampling.
Each ResNet block employs $3 \times 3$ convolutions with group normalization \citep{wu2018group} (8 groups, $\epsilon = 10^{-6}$) and GELU activation functions \citep{hendrycks2023gaussian}.
At the bottleneck, we incorporate multi-head self-attention with 4 heads to capture long-range spatial dependencies before projecting to the latent space.

The latent space models a diagonal Gaussian distribution with separate learned mean and log-variance channels.
We clamp the log-variance to $[-30, 20]$ for numerical stability and use the reparameterization trick during training to enable backpropagation through the stochastic sampling process.
The decoder mirrors the encoder architecture with transposed convolutions for upsampling, progressively reconstructing the full spectral resolution through channel dimensions $[128, 256, 512]$ before a final projection to 1028 output channels.

\subsection{Training Procedure}

We optimize the VAE using a weighted combination of reconstruction and KL divergence losses.
The reconstruction loss employs L1 distance between input and output: $\mathcal{L}_{\mathrm{rec}} = ||x - \hat{x}||_1$.
We use a learnable variance parameter $\log\sigma^2$ (initialized to 6.0) to adaptively weight the reconstruction loss: $\mathcal{L}_{\mathrm{nll}} = \mathcal{L}_{\mathrm{rec}} / \exp(\log\sigma^2) + \log\sigma^2$.
The KL divergence loss regularizes the latent distribution toward a standard Gaussian: $\mathcal{L}_{\mathrm{KL}} = -0.5 \sum(1 + \log\sigma_z^2 - \mu_z^2 - \sigma_z^2)$.
The total loss combines these with a small KL weight: $\mathcal{L}_{\mathrm{total}} = \mathcal{L}_{\mathrm{nll}} + 10^{-6} \cdot \mathcal{L}_{\mathrm{KL}}$, prioritizing reconstruction quality while maintaining regularization.

Training uses the AdamW optimizer with learning rate $10^{-4}$, weight decay $0.05$, and gradient clipping (max norm 1.0).
The model trains for 200,000 steps with batch size 32 on a single GPU (see Appendix~\ref{app:model_training} for additional details).

\subsection{Component Extraction via Probing}

To evaluate whether the learned latent representations encode atmospheric information, we employ a two-stage probing approach.
After training the VAE, we freeze its weights and train supervised probes from the latent space to Level-2 atmospheric products.
We extract the mean of the latent distribution (32 channels at $16\times16$ spatial resolution) as input features for the probes.

For linear probing, we train a single linear layer mapping from 32 latent channels to 1 output channel for each atmospheric component.
For nonlinear probing, we employ a three-layer MLP with architecture $[32 \rightarrow 512 \rightarrow 512 \rightarrow 1]$, using ReLU activations and dropout (0.1) between layers.
Both probe types are optimized with AdamW using MSE loss (see Appendix~\ref{app:probing} for training details).

\begin{figure}[ht]
    \centering
    \includegraphics[width=0.95\columnwidth]{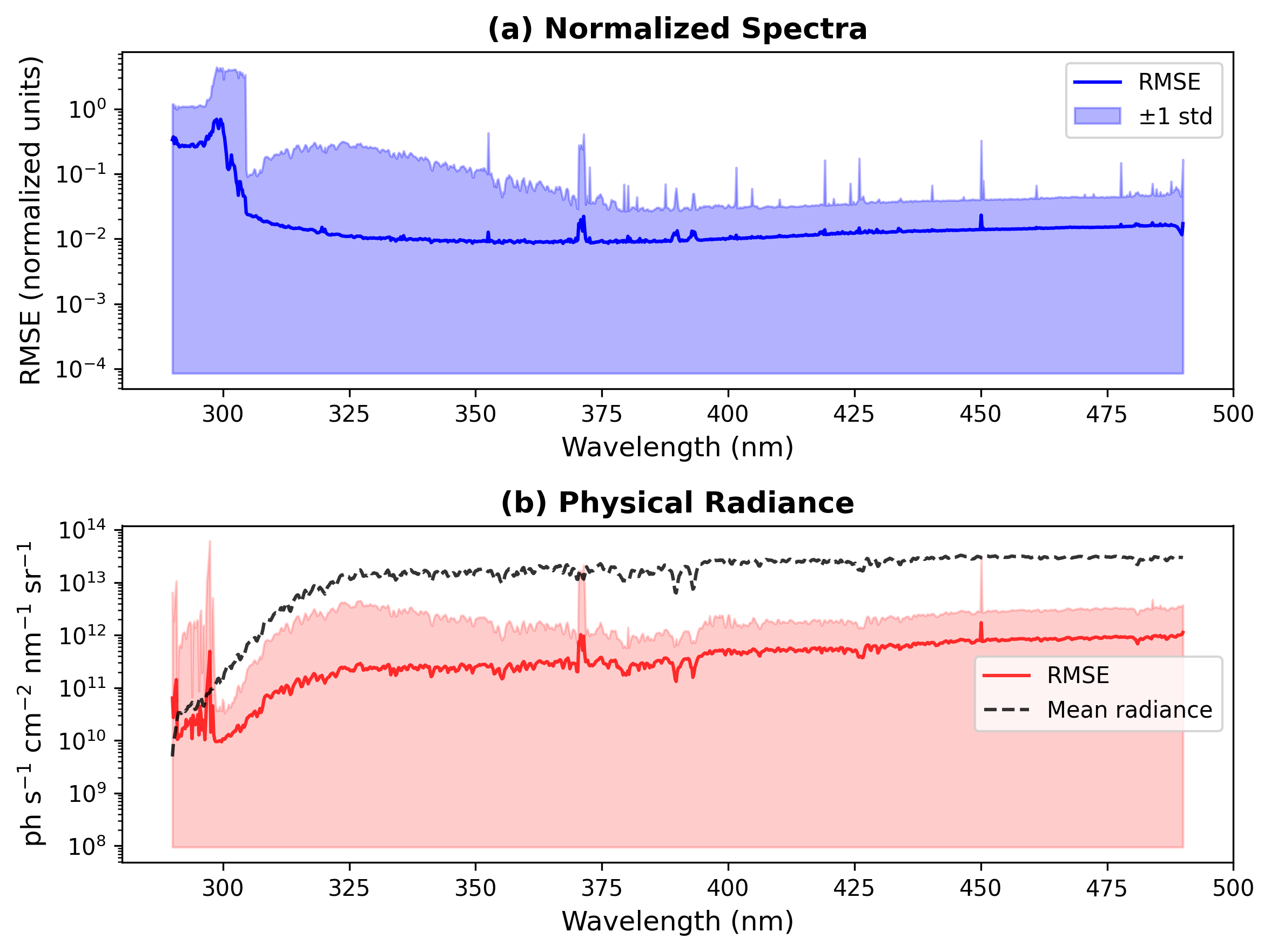}
    \caption{\textbf{Channel-wise reconstruction errors across validation set.} Root mean squared error (RMSE) computed per spectral channel over 61,440 validation spectra sampled from 960 tiles. (a) RMSE for normalized spectra showing wavelength-dependent reconstruction quality with errors ranging from $10^{-2}$ to $10^{-1}$ normalized units. (b) RMSE for physical radiance (red line) overlaid with mean radiance spectrum (black dashed line), demonstrating that reconstruction errors are 1-2 orders of magnitude smaller than the signal across the 290-490nm range, with slightly elevated errors only at the shortest wavelengths (290-310nm). Shaded regions indicate ±1 standard deviation.}
    \label{fig:quantitative_reconstruction}
\end{figure}

\section{Results}

\begin{figure*}[ht]
\centering
\includegraphics[width=0.49\textwidth]{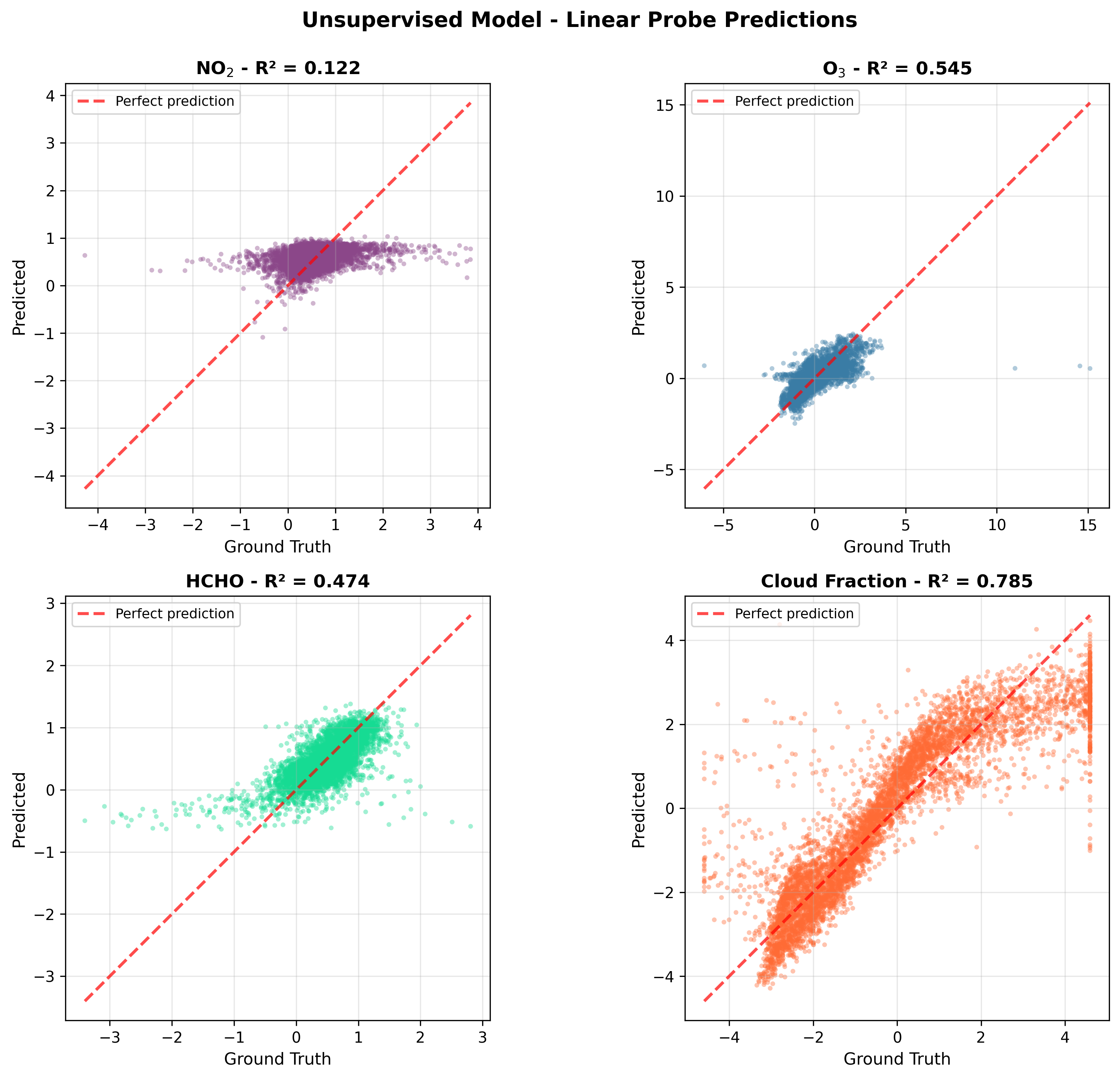}
\includegraphics[width=0.49\textwidth]{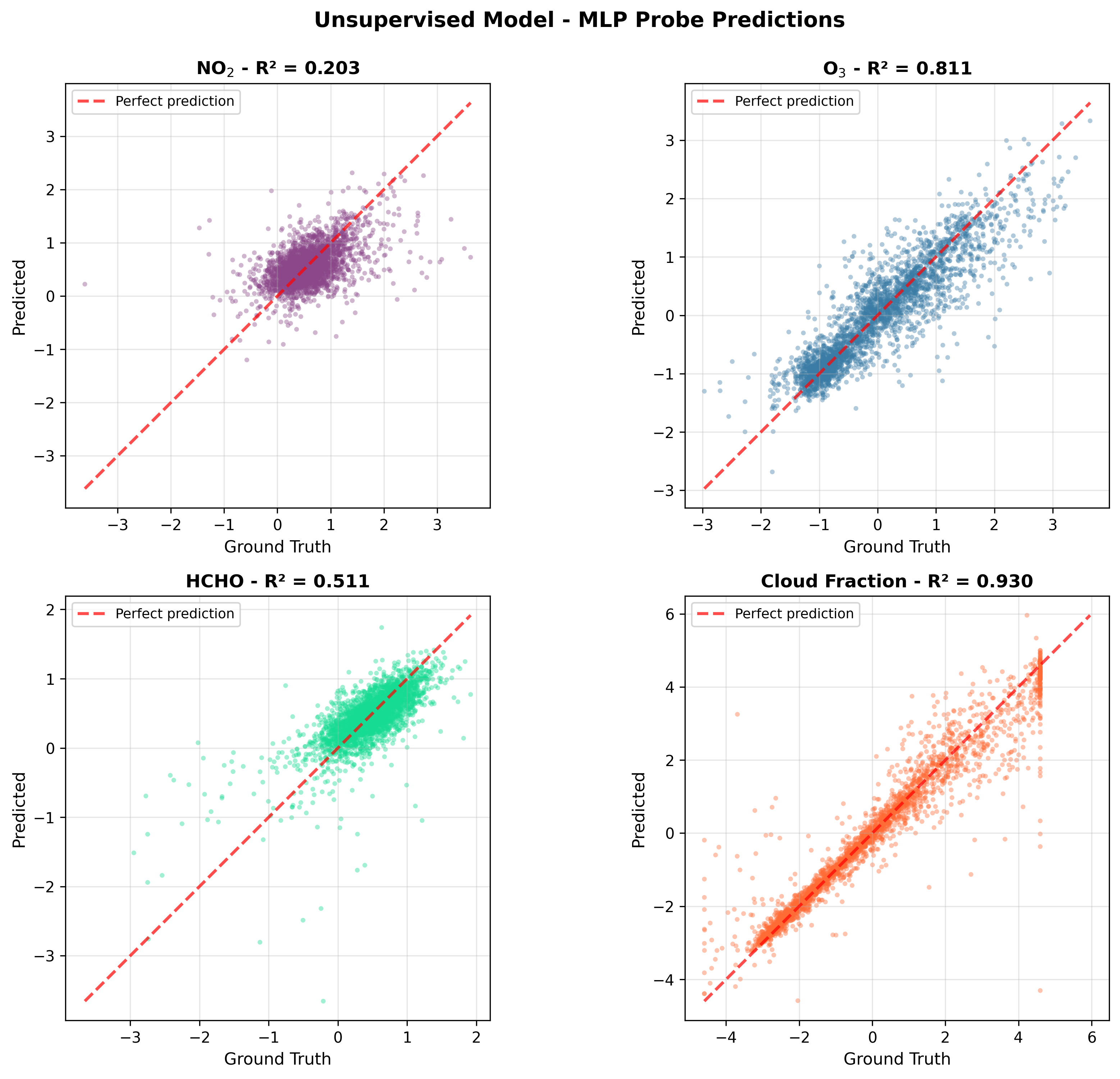}
\caption{\textbf{Unsupervised VAE probing results.} Predicted vs. ground truth scatter plots for all four atmospheric products extracted from the base VAE latent space using (left) linear probes and (right) MLP probes. Each 2×2 grid shows: (a) NO$_2$ tropospheric vertical column, (b) total O$_3$ column, (c) HCHO vertical column, (d) cloud fraction. Red dashed lines indicate perfect prediction. MLP probes substantially outperform linear probes, particularly for cloud fraction (R$^2$: 0.785$\rightarrow$0.930) and total ozone (R$^2$: 0.545$\rightarrow$0.811), demonstrating that atmospheric information is encoded nonlinearly in the unsupervised latent representation. NO$_2$ remains challenging (R$^2$=0.203) despite the nonlinear extraction, indicating fundamental limitations of purely unsupervised compression for this trace gas.}
\label{fig:probing_results}
\end{figure*}

We evaluate our VAE-based compression approach on TEMPO hyperspectral data through reconstruction quality metrics and the ability to extract atmospheric products from the compressed latent representations.

\subsection{Compression and Reconstruction Performance}

Our VAE achieves a compression ratio of $\times$514, reducing the data from $[1028 \times 64 \times 64]$ to $[32 \times 16 \times 16]$ latent dimensions. Despite this aggressive compression, reconstruction errors remain very low across nearly all spectral channels and spatial locations. Figure \ref{fig:quantitative_reconstruction} shows channel-wise reconstruction errors across the full validation set: errors are consistently 1-2 orders of magnitude below the signal across the 290-490nm range, with slightly elevated errors only at the shortest wavelengths (290-310nm). Figure \ref{fig:reconstruction} demonstrates representative examples showing close agreement between original and reconstructed radiance spectra at individual spatial locations, with difference maps confirming the low error magnitudes.

\subsection{Unsupervised Component Extraction}

We evaluate the extraction of four key atmospheric products from the compressed representations using both linear and MLP probes. Figure \ref{fig:probing_results} presents scatter plots comparing predicted versus ground truth values for all four products.

The results demonstrate strong performance for cloud fraction (R$^2$=0.930) and total ozone (R$^2$=0.811) extraction using MLP probes. Cloud detection benefits from strong spatial coherence, with the convolutional architecture capturing extended cloud patterns to achieve near-perfect reconstruction. Total ozone also performs well despite the aggressive compression, with MLP probes improving substantially over linear (R$^2$: 0.545→0.811), demonstrating that the VAE latent space successfully encodes the broad stratospheric ozone distributions visible in Figure \ref{fig:tempo_data}.

NO$_2$ extraction remains challenging with MLP probes achieving R$^2$=0.203 compared to R$^2$=0.122 for linear probes, representing a 67\% improvement but still indicating substantial room for improvement. This difficulty stems from NO$_2$'s inherent challenges: low signal-to-noise ratio, complex dependence on surface properties and atmospheric transport, and—as visible in Figure \ref{fig:tempo_data}—noise-like spatial structure lacking the extended coherence that aids compression of other products like clouds and ozone. HCHO shows moderate performance with MLP (R$^2$=0.511) providing minimal improvement over linear probes (R$^2$=0.474), suggesting its signal is partially but incompletely captured.

The comparison between linear and MLP probes reveals that the VAE encodes atmospheric information in a semi-linear manner. MLP probes significantly outperform linear probes for cloud fraction (R$^2$: 0.785→0.930) and O$_3$ (0.545→0.811), while NO$_2$ shows substantial improvement (0.122→0.203) and HCHO minimal change (0.474→0.511). This pattern indicates that atmospheric products are not directly accessible through linear combinations of latent dimensions—nonlinear transformations are necessary to extract the encoded information, with varying degrees of difficulty across products.

A natural question arises: the unsupervised VAE has no explicit objective to preserve atmospheric product information—it only minimizes reconstruction error. Would directly supervising the latent space to predict L2 products during training improve component extraction?

\begin{figure}[h]
    \centering
    \includegraphics[width=0.8\columnwidth]{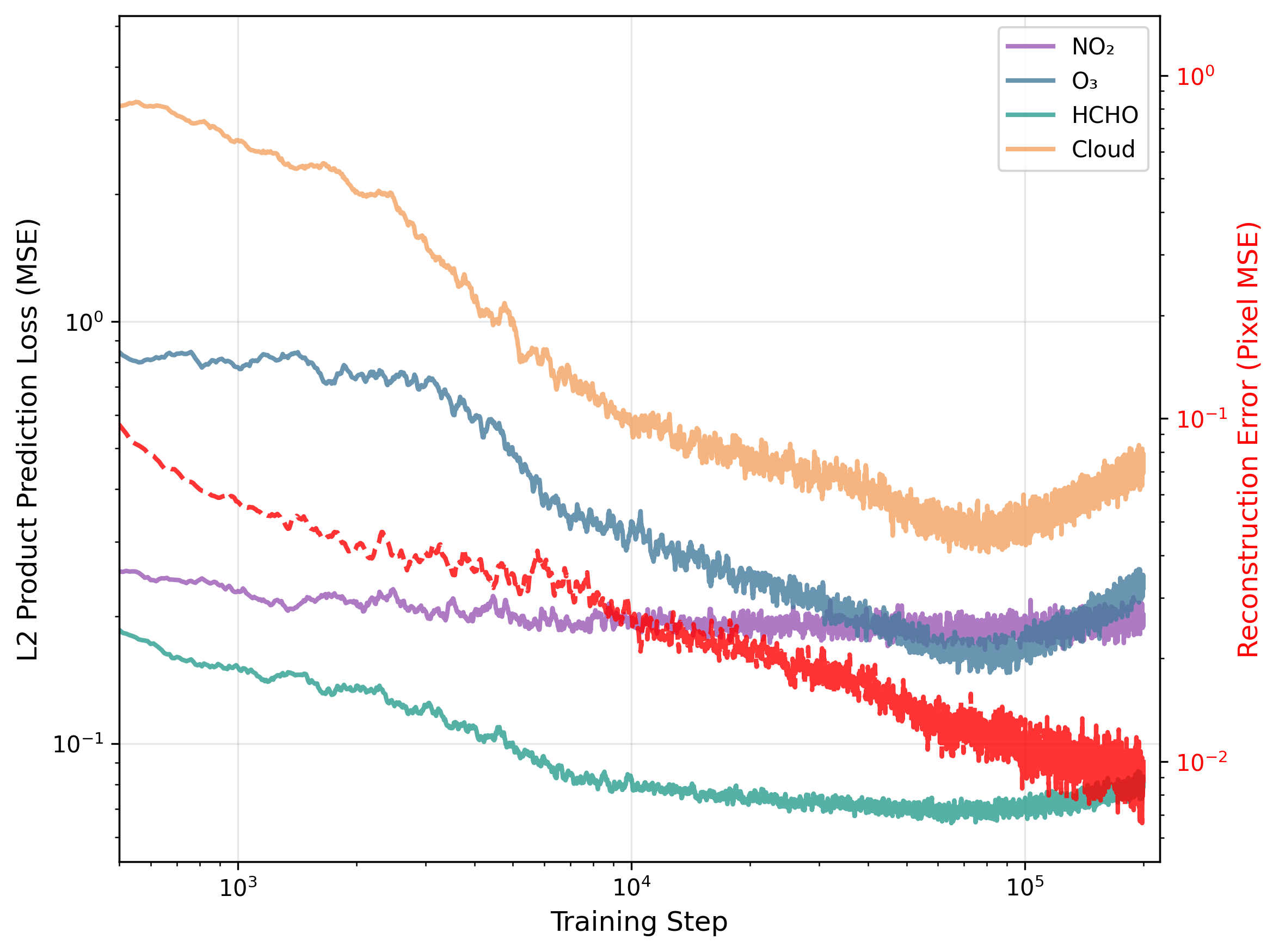}
    \caption{\textbf{Joint optimization of reconstruction and L2 prediction.} Combined view of L2 product prediction losses (left y-axis) and reconstruction error (right y-axis, red dashed line) during latent supervised VAE training over 220,000 steps. L2 product losses (NO$_2$ in purple, O$_3$ in blue, HCHO in teal, cloud in orange) improve significantly once reconstruction error plateaus around step 40,000. Cloud, O$_3$, and HCHO show overfitting in later stages, while NO$_2$ barely improves throughout training, indicating fundamental encoding difficulties for this trace gas even under explicit supervision.}
    \label{fig:supervised_training}
    \end{figure}

\begin{figure*}[h!]
    \centering
    \includegraphics[width=0.49\textwidth]{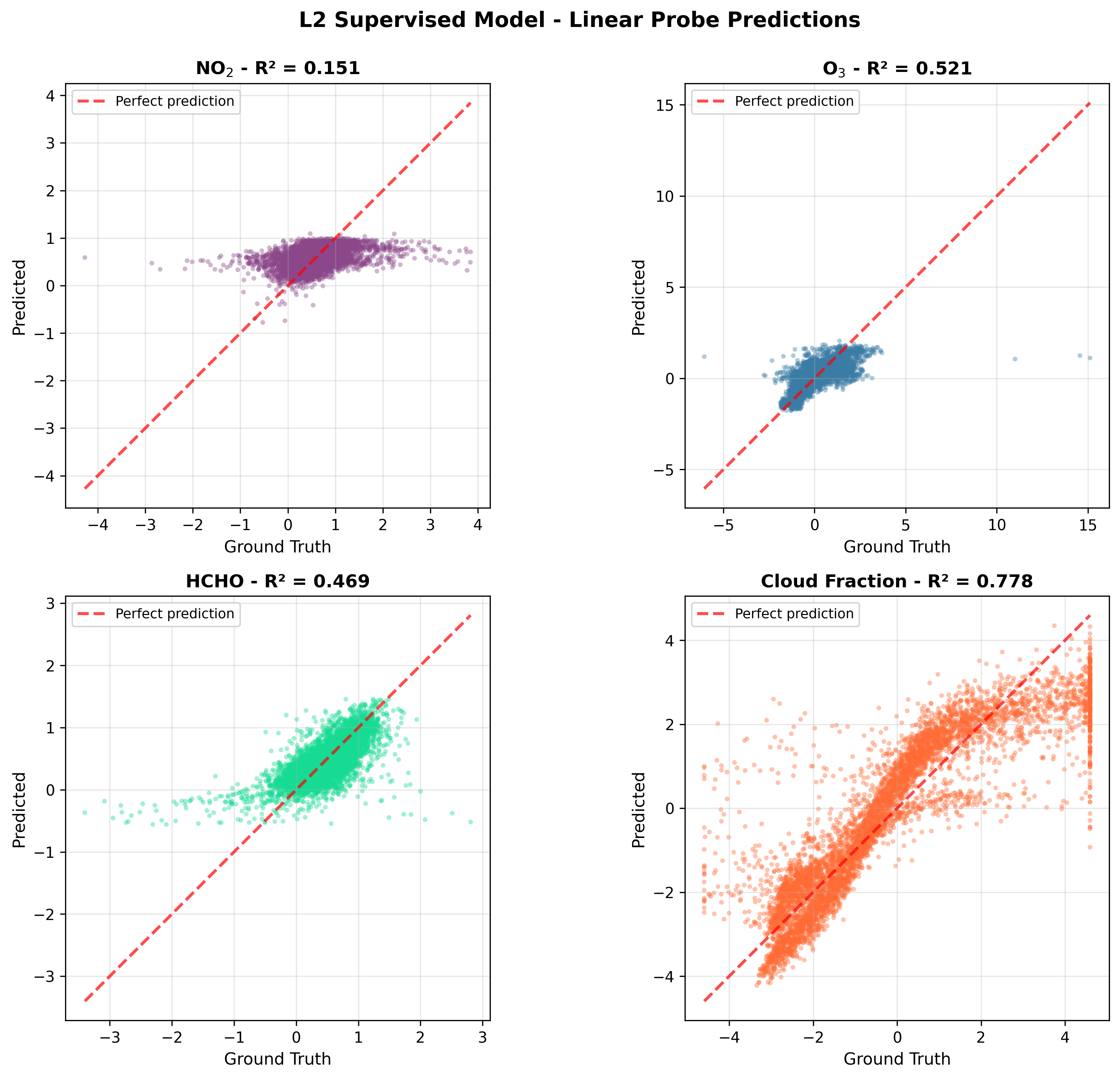}
    \includegraphics[width=0.49\textwidth]{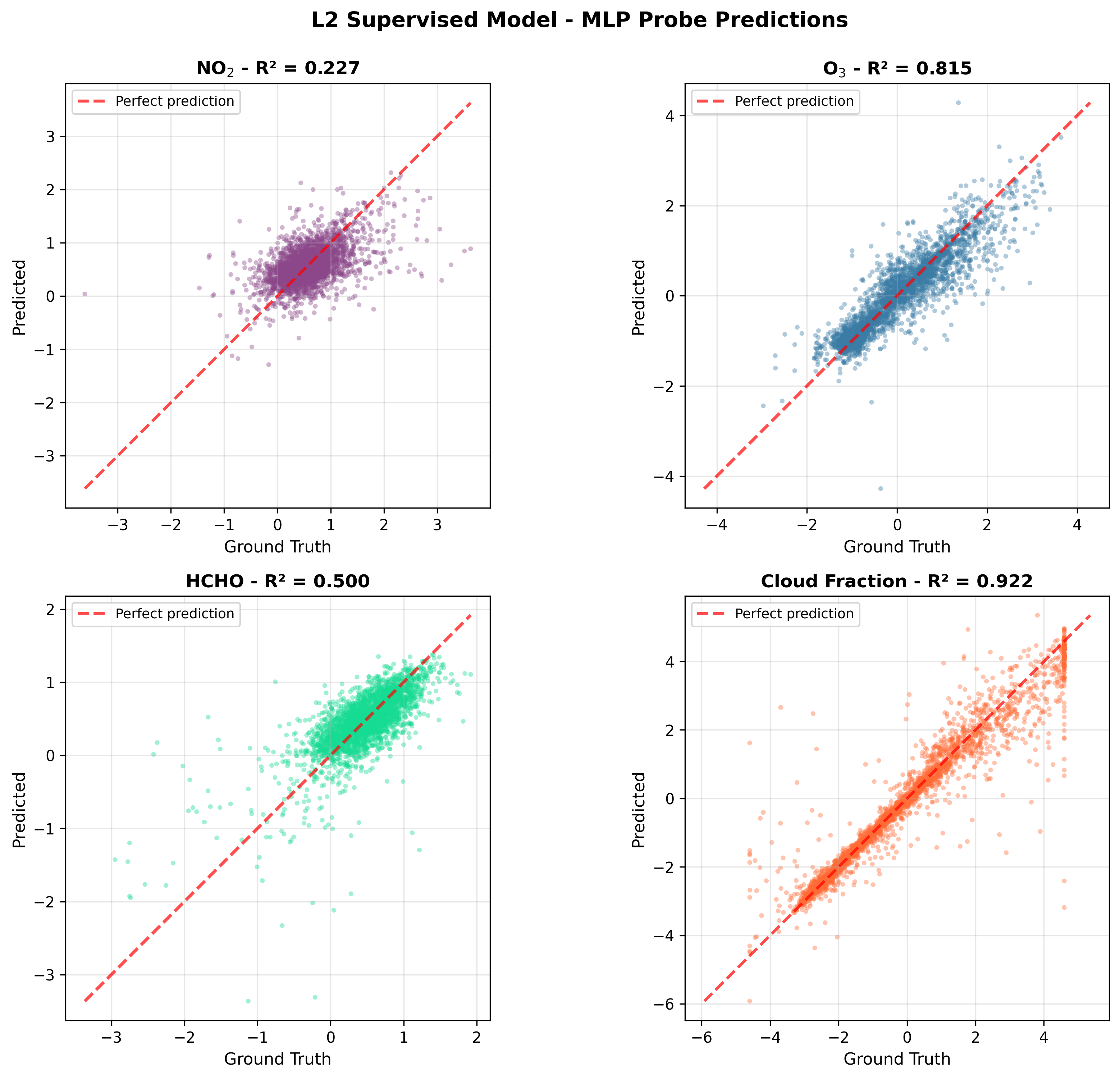}
    \caption{\textbf{L2-supervised VAE probing results.} Predicted vs. ground truth scatter plots for atmospheric products extracted from the L2-supervised VAE latent space using (left) linear probes and (right) MLP probes. Despite explicit supervision during VAE training to predict L2 products from latent representations, the probing results show minimal improvement compared to the unsupervised model (Figure \ref{fig:probing_results}). MLP probes again outperform linear probes with similar margins: cloud fraction achieves R$^2$=0.922, total ozone R$^2$=0.815, while NO$_2$ shows modest improvement to R$^2$=0.227. This similarity suggests that the VAE's reconstruction objective already captures sufficient atmospheric structure, with explicit supervision providing only marginal benefits for post-hoc product extraction.}
    \label{fig:supervised_probing}
    \end{figure*}

\subsection{Latent Supervised Component Extraction}

To test this, we trained a supervised VAE variant that jointly optimizes reconstruction loss and explicit L2 product prediction from the latent representation. Figure \ref{fig:supervised_training} shows the training dynamics of this multi-task learning approach. All L2 product prediction losses decrease steadily throughout training, with the latent readout performance improving significantly once the reconstruction loss plateaus around step 40,000. Cloud fraction, O$_3$, and HCHO show signs of overfitting in later training stages with increasing divergence between training and validation curves, while NO$_2$ shows minimal improvement throughout, barely decreasing from its initial loss value. This demonstrates successful joint optimization for some products but limited progress for NO$_2$, suggesting fundamental challenges in encoding this trace gas even with explicit supervision. However, as shown in Figure \ref{fig:supervised_probing}, despite this explicit supervision, the supervised model achieves nearly identical probe performance compared to the unsupervised baseline.

To investigate whether explicit supervision during training improves atmospheric product extraction, we trained a supervised VAE variant that jointly optimizes reconstruction loss and L2 product prediction from the latent space. Surprisingly, the supervised model achieves nearly identical probe performance compared to the unsupervised baseline (Figure \ref{fig:supervised_probing}). Cloud fraction (R$^2$=0.922 vs. 0.923), total ozone (R$^2$=0.815 vs. 0.754), HCHO (R$^2$=0.500 vs. 0.498), and NO$_2$ (R$^2$=0.227 vs. 0.209) show minimal differences. This result reveals that the difficulty in extracting certain atmospheric components stems from fundamental encoding challenges rather than lack of supervision—the reconstruction objective already captures sufficient atmospheric structure, and explicit supervision during training provides negligible benefit for subsequent product extraction from the compressed representation.

The unexpected difficulty in extracting NO2 compared to HCHO (R²=0.20 vs 0.51) despite NO2's stronger spectral signal warrants further investigation. This counterintuitive result suggests that factors beyond signal strength—such as spatial variability, interference from other absorbers, or the complexity of NO2's fine-scale urban patterns—may dominate the extraction challenge.

Importantly, our current approach targets Level-2 products, which incorporate radiative transfer corrections and ancillary data beyond the pure spectral information. Future work could target slant column densities (the direct spectral retrieval before atmospheric corrections) to isolate the spectral information content from uncertainties introduced during the conversion to vertical columns. This approach would bypass the substantial ancillary data dependencies and potential errors in the air mass factor calculations, potentially revealing stronger correlations between the compressed spectral representations and the atmospheric constituents.

\section{Conclusion}

We present a variational autoencoder approach for joint spatial-spectral compression of TEMPO hyperspectral satellite data, achieving $\times$514 compression with excellent spectral reconstruction fidelity—errors are 1-2 orders of magnitude below the signal across all 1028 channels. This dramatic data volume reduction addresses critical challenges in storing, transmitting, and sharing the terabytes of hyperspectral observations generated daily by geostationary satellites. Beyond compression, our results demonstrate that unsupervised representation learning preserves atmospheric information, enabling extraction of certain products like cloud fraction (R$^2$=0.923) and total ozone (R$^2$=0.754) directly from compressed representations.

The strong performance on cloud and ozone detection validates the potential of learned compression for satellite data analysis. These products, which have clear spectral signatures and spatial patterns, are well-preserved even under aggressive compression. Importantly, we find that atmospheric information is encoded in a semi-linear manner—while linear probes extract some information, nonlinear MLP probes substantially outperform them (e.g., O$_3$: R$^2$ 0.545→0.754), indicating that atmospheric products require nonlinear transformations for extraction from the compressed latent space.

However, tropospheric trace gases NO$_2$ and HCHO remain challenging (R$^2$=0.21 and 0.50 respectively), reflecting both the inherent difficulty of these retrievals and limitations of purely unsupervised compression. These products have weak spectral signatures, low atmospheric concentrations, and complex dependencies on surface properties and atmospheric transport, making them fundamentally challenging for any retrieval approach. Critically, our supervised VAE experiment—which explicitly optimizes for L2 product prediction during training—yields nearly identical probe performance to the unsupervised baseline, revealing that the difficulty stems from fundamental encoding challenges rather than lack of supervision. This suggests that improving trace gas extraction may require fundamentally different architectural approaches or explicit physical constraints beyond standard supervised learning.

Future work will explore multi-task learning frameworks that jointly optimize reconstruction and product extraction objectives, potentially improving performance on challenging retrievals while maintaining compression efficiency. Additionally, incorporating physical constraints and temporal modeling could further enhance the representation quality. 

The inclusion of TEMPO's solar irradiance reference spectra, measured weekly by direct solar observation, presents another promising direction. These measurements provide a clean spectral baseline without atmospheric absorbers, which could be leveraged in several ways: (1) as an additional input channel to help the VAE learn to separate atmospheric from instrumental signatures, (2) as a normalization reference to create absorption spectra before encoding, or (3) as a regularization target to ensure the latent space preserves physically meaningful spectral ratios. Such physics-informed approaches could particularly benefit the extraction of weak absorbers like NO$_2$ and HCHO.

The approach presented here establishes a foundation for applying deep learning to hyperspectral satellite data compression and analysis, with implications for next-generation Earth observation systems facing increasing data volumes.

\subsubsection*{Acknowledgments}
CFP gratefully acknowledges the support of Hidenori Tanaka, Douglas Finkbeiner and Aravinthan D.T. Samuel. CFP thanks the Kempner Institute for the Study of Natural and Artificial Intelligence at Harvard University for academic support. The computations in this paper were run on the FASRC cluster supported by the FAS Division of Science Research Computing Group at Harvard University.

\bibliography{aaai2026}

\appendix
\setcounter{secnumdepth}{2}
\renewcommand{\thesection}{\Alph{section}}
\renewcommand{\thesubsection}{\thesection.\arabic{subsection}}

\vspace{1em}
\begin{center}
\textbf{\Large APPENDIX}
\end{center}
\vspace{1em}

\section{Data}
\label{app:data}

TEMPO (Tropospheric Emissions: Monitoring of Pollution) is NASA's first Earth Venture-Instrument mission designed to monitor atmospheric pollution across North America from geostationary orbit. Launched on April 7, 2023, aboard the commercial satellite Intelsat-40e (IS-40e), TEMPO operates at approximately 22,000 miles above Earth's equator at 91°W longitude. This vantage point provides a constant view of North America, enabling the instrument to scan the continent from the Atlantic to the Pacific and from the Yucatán Peninsula to northern Canada with unprecedented temporal and spatial resolution. TEMPO is part of a global constellation of geostationary air quality monitoring satellites, alongside Sentinel-4 (Europe) and GEMS (Asia).

The TEMPO instrument is a grating spectrometer sensitive to ultraviolet and visible wavelengths (290-740 nm, split across two spectral bands: 290-490 nm and 540-740 nm). The instrument employs a scanning mirror that steps East to West across the field of regard, collecting approximately 2.4 million spectra per hour during nominal hourly scans. Each scan captures reflected sunlight from Earth's surface and atmosphere using two 2D CCD detectors (2048 spatial pixels $\times$ 1028 spectral channels for the UV-visible band, 290-490 nm). The spatial resolution is approximately 2 km North-South $\times$ 4.75 km East-West at the center of the field of regard, enabling detection of air quality variations at sub-urban scales. TEMPO's primary data products include Level-1B calibrated radiances and Level-2 atmospheric products derived from spectral retrievals: nitrogen dioxide (NO$_2$), ozone (O$_3$), formaldehyde (HCHO), aerosol optical depth, sulfur dioxide (SO$_2$), and cloud properties.

For this work, we utilize TEMPO Level-1B radiance data and corresponding Level-2 products from January 2025 covering the Los Angeles region. Our dataset consists of 70 granules spanning various times of day and atmospheric conditions. We focus on the UV-visible spectral band (290-490 nm, 1028 channels) which contains the primary absorption features for trace gas retrievals. From Level-2 products, we extract tropospheric NO$_2$ vertical columns, total ozone columns, HCHO vertical columns, and cloud fraction to evaluate how well atmospheric information is preserved in compressed representations. All data are obtained from NASA's Earthdata portal in netCDF4 format.

\subsection{Normalization Strategy}
\label{app:data_norm}

Proper normalization of TEMPO data is critical for effective neural network training, as the raw measurements span multiple orders of magnitude and exhibit strong wavelength-dependent variations. We employ distinct normalization strategies for Level-1B radiances and Level-2 atmospheric products, tailored to the statistical properties of each data type.

\textbf{Level-1B Radiance Normalization.} Raw TEMPO radiances exhibit exponentially distributed values with significant dynamic range. We apply a three-stage normalization pipeline: (1) log transformation to compress the dynamic range: $\log(\max(\mathrm{radiance}, 1.0))$, where the clamping threshold of 1.0 prevents numerical issues with near-zero values, (2) channel-wise z-score normalization using global statistics computed from 10 representative files containing 2,414,592 total pixels: $z = (\log(r) - \mu_\lambda) / (\sigma_\lambda + 10^{-8})$, where $\mu_\lambda$ and $\sigma_\lambda$ are the mean and standard deviation for spectral channel $\lambda$, and (3) clipping to $[-10, 10]$ to remove outliers while preserving $>99.9\%$ of the data distribution. The per-channel normalization preserves spectral structure while accounting for wavelength-dependent sensor sensitivities and atmospheric transmittance. Table \ref{tab:normalization} shows the computed global statistics span mean values from 4.80 to 30.68 (in log-radiance space) and standard deviations from 1.13 to 12.18 across the 1028 spectral channels, reflecting the strong wavelength dependence of Earth's reflectance spectrum.

\begin{table}[h]
\centering
\caption{Global normalization statistics computed from 10 TEMPO files (2,414,592 pixels) showing per-channel statistics in log-radiance space.}
\small
\begin{tabular}{lcc}
\hline
\textbf{Statistic} & \textbf{Min} & \textbf{Max} \\
\hline
Mean ($\mu_\lambda$) & 4.796 & 30.683 \\
Std Dev ($\sigma_\lambda$) & 1.128 & 12.185 \\
\hline
\end{tabular}
\label{tab:normalization}
\end{table}

\textbf{Level-2 Product Normalization.} Each atmospheric product requires specialized normalization to handle its unique statistical properties and physical constraints. Raw L2 values are first converted to convenient units by scaling: NO$_2$ and HCHO vertical columns (originally in molec/cm$^2$) are divided by $10^{15}$ and $10^{16}$ respectively to bring values to O(1), while total ozone (already in Dobson Units) and cloud fraction (already in [0,1]) require no scaling. We then apply product-specific transformations: (1) \textbf{NO$_2$ and HCHO}: Inverse hyperbolic sine (asinh) transformation with robust MAD-based scaling to handle negative values and heavy-tailed distributions common in trace gas measurements: $\mathrm{asinh}(x / s)$ where $s = 1.4826 \times \mathrm{MAD}$, computed from the median absolute deviation of the training data. This transformation provides symmetric treatment of positive and negative values while compressing extreme outliers. (2) \textbf{Total Ozone}: Standard z-score normalization $(x - \mu) / \sigma$ computed from training data, as ozone columns exhibit approximately Gaussian distributions with moderate variance. (3) \textbf{Cloud Fraction}: Logit transformation with boundary squeeze to spread the $[0,1]$-bounded values across the real line: $\mathrm{logit}(\epsilon + (1-2\epsilon) \cdot x)$ where $\epsilon=0.01$ prevents infinities at the boundaries. This transformation is particularly effective for quantities with strong boundary effects. Table \ref{tab:l2products} summarizes the normalization specifications for each product.

\begin{figure}[h]
\centering
\includegraphics[width=\columnwidth]{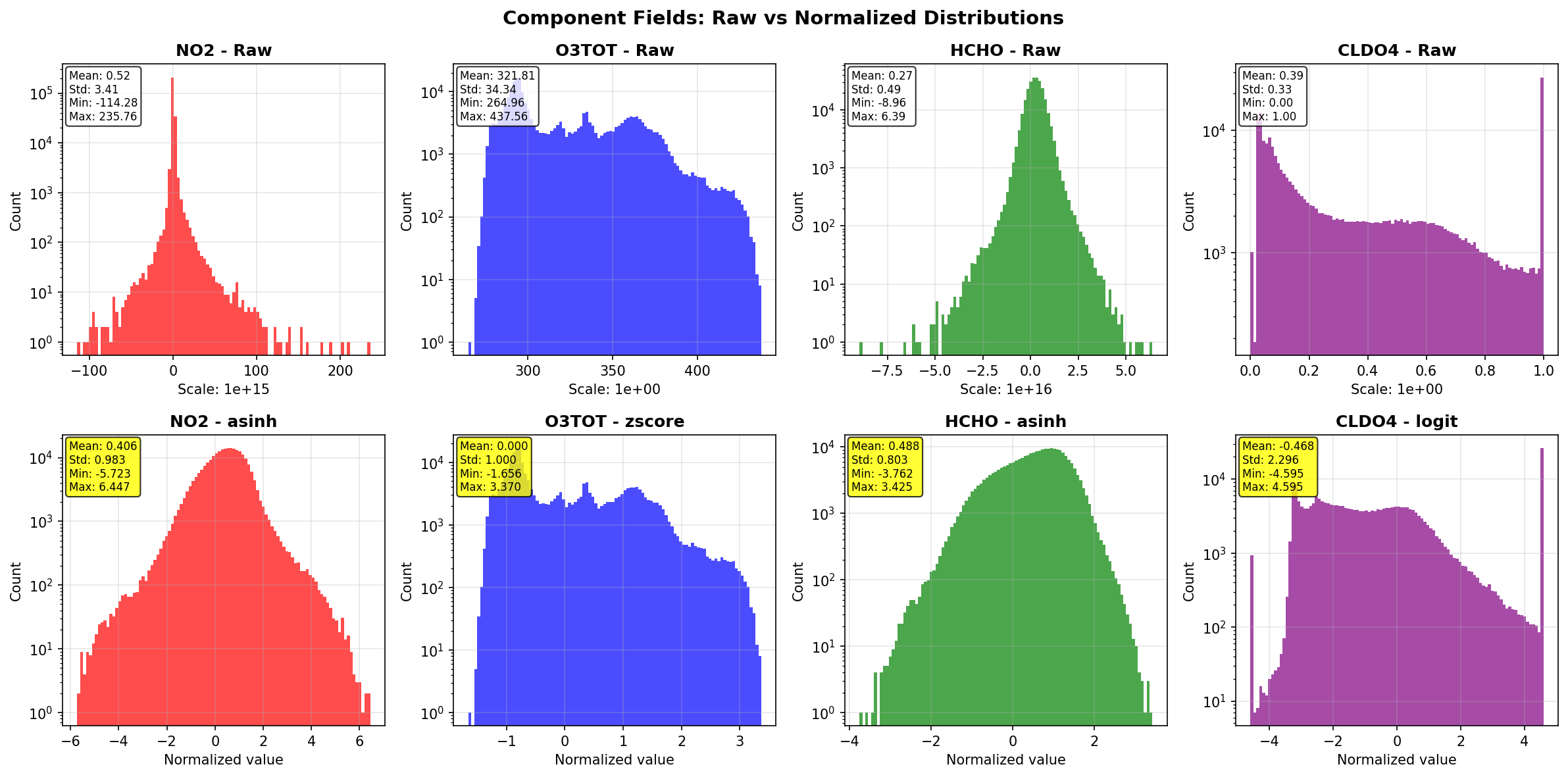}
\caption{\textbf{Normalization strategy comparison.} Each row shows a different normalization approach applied to TEMPO hyperspectral data: (top) global statistics across all channels and pixels, (middle) per-channel normalization, and (bottom) hybrid approach. Left column displays the normalized radiance spectra, while right column shows the spatial distribution of a representative spectral channel. The hybrid approach adopted in our work preserves both spectral signatures and spatial patterns necessary for atmospheric retrieval.}
\label{fig:normalization}
\end{figure}

Figure \ref{fig:normalization} illustrates the impact of different normalization strategies on TEMPO spectral data. The comparison demonstrates how various preprocessing approaches affect the representation of radiance measurements across the 290-490nm wavelength range. The global per-channel normalization approach (adopted in this work) preserves relative spectral intensities across spatial locations while accounting for wavelength-dependent variations, maintaining both spectral structure and spatial variability essential for accurate atmospheric product retrieval.

For Level-2 atmospheric products, we apply product-specific normalization strategies shown in Table \ref{tab:l2products}.

\begin{table}[h]
\centering
\caption{Processing specifications for atmospheric products.}
\small
\begin{tabular}{llll}
\hline
\textbf{Product} & \textbf{Units} & \textbf{Scale} & \textbf{Norm.} \\
\hline
NO$_2$ & molec/cm$^2$ & $10^{15}$ & asinh \\
O$_3$ & DU & 1.0 & z-score \\
HCHO & molec/cm$^2$ & $10^{16}$ & asinh \\
Cloud & fraction & 1.0 & logit \\
\hline
\end{tabular}
\label{tab:l2products}
\end{table}

For asinh normalization, we use: $\mathrm{asinh}(x / s)$ where $s = 1.4826 \times \mathrm{MAD}$ (Median Absolute Deviation).
For logit normalization, we apply: $\mathrm{logit}(0.01 + 0.98 \times x)$ to avoid infinities at boundaries.

\section{Model Specification}
\label{app:model}

Our approach employs three types of neural network models with distinct roles in the pipeline. The primary model is a Variational Autoencoder (VAE) that learns to compress TEMPO's high-dimensional radiance data from $[1028 \times 64 \times 64]$ input tensors to compact $[32 \times 16 \times 16]$ latent representations through purely unsupervised learning on Level-1B data. This VAE consists of a convolutional encoder, a probabilistic latent space with reparameterization, and a symmetric decoder for reconstruction. After VAE training, we freeze the encoder and evaluate whether atmospheric information is preserved in the latent space by training two types of supervised probes: (1) linear probes, which are single-layer linear mappings from latent channels to atmospheric products, and (2) MLP probes, which are three-layer feedforward networks with ReLU activations and dropout. The comparison between linear and MLP probe performance reveals whether atmospheric signals are encoded linearly or require nonlinear extraction from the compressed representation.

To investigate whether explicit supervision improves atmospheric product extraction, we also train a latent supervised VAE model with a modified architecture that includes prediction heads for the four L2 atmospheric products (NO$_2$, O$_3$, HCHO, cloud fraction). This latent supervised VAE uses a multi-task learning objective that combines the standard VAE reconstruction and KL divergence losses with supervised prediction losses on the L2 products. The latent space is jointly optimized for both reconstruction fidelity and atmospheric product predictability, encouraging the learned representations to explicitly encode the target atmospheric signals. This approach allows us to assess whether incorporating supervision during training improves the accessibility of atmospheric information compared to purely unsupervised compression followed by post-hoc probing.

\subsection{VAE Architecture Details}
\label{app:model_arch}

\textbf{Encoder.} The encoder transforms input tensors of shape $[1028 \times 64 \times 64]$ through hierarchical spatial compression. An initial $3 \times 3$ convolution projects the 1028 spectral channels to 512 internal channels. The encoder then consists of three downsampling levels with channel dimensions $[512, 256, 128]$, where each level contains one ResNetBlock followed by strided convolution (kernel=2, stride=2) for $2 \times$ spatial downsampling. This produces intermediate representations at resolutions $32 \times 32$ (512 channels), $16 \times 16$ (256 channels), and $8 \times 8$ (128 channels). Each ResNetBlock contains two convolutional paths: (1) GroupNorm (8 groups, eps=$10^{-6}$) → GELU → $3 \times 3$ Conv, followed by (2) GroupNorm → GELU → $3 \times 3$ Conv with zero-initialized weights. A $1 \times 1$ skip connection adapts channel dimensions when input and output differ. After spatial downsampling, a middle block at $8 \times 8$ resolution applies two ResNetBlocks with a multi-head self-attention layer (4 heads, 32 channels per head) between them. The attention mechanism uses full spatial self-attention across all 64 spatial positions with scaled dot-product: $\text{Attention}(Q,K,V) = \text{softmax}(QK^T/\sqrt{d_k})V$. Finally, a $3 \times 3$ convolution with zero initialization projects to 64 channels encoding mean and log-variance: the final latent distribution has shape $[32 \times 8 \times 8]$ for both parameters.

\textbf{Latent Space.} We parameterize a diagonal Gaussian distribution with separate learned mean $\mu$ and log-variance $\log\sigma^2$ for each of the $32 \times 8 \times 8 = 2048$ latent variables. Log-variance is clamped to [-30, 20] for numerical stability. Sampling uses the reparameterization trick: $z = \mu + \sigma \epsilon$ where $\epsilon \sim \mathcal{N}(0,I)$ and $\sigma = \exp(0.5\log\sigma^2)$. The KL divergence against a standard Gaussian prior $\mathcal{N}(0,I)$ is computed as $D_{KL} = 0.5 \sum (\mu^2 + \sigma^2 - 1 - \log\sigma^2)$.

\textbf{Decoder.} The decoder mirrors the encoder architecture with symmetric structure. Starting from latent shape $[32 \times 8 \times 8]$, an initial projection expands to 128 channels, followed by an identical middle block (two ResNetBlocks with 4-head attention between them). Three upsampling levels progressively increase spatial resolution using transposed convolutions (kernel=2, stride=2): $8 \times 8 \rightarrow 16 \times 16$ (128→256 channels), $16 \times 16 \rightarrow 32 \times 32$ (256→512 channels), and one final ResNetBlock at $32 \times 32 \rightarrow 64 \times 64$ (512 channels maintained). Each level contains one ResNetBlock followed by channel expansion via transposed convolution, except the final level which omits upsampling to preserve the target $64 \times 64$ resolution. A final $3 \times 3$ convolution with zero initialization projects from 512 internal channels back to 1028 output spectral channels.

\textbf{Loss Function.} The total loss combines reconstruction and regularization: $\mathcal{L} = ||x - \hat{x}||_1 / \exp(\log s^2) + \log s^2 + 10^{-6} \cdot D_{KL}$, where the reconstruction term uses L1 distance weighted by a learned log-variance $\log s^2$ (initialized to 6.0), and the KL divergence is weighted by $10^{-6}$.

\textbf{Key Design Choices.} (1) Attention is applied only at the $8 \times 8$ bottleneck resolution. (2) Heavy channel usage (512 channels at finest resolution). (3) Zero initialization of residual path outputs and final projections. (4) No dropout (probability=0.0). The complete model contains approximately 54 million parameters.

\subsection{Training Configuration}
\label{app:model_training}

\begin{figure}[ht]
\centering
\includegraphics[width=\columnwidth]{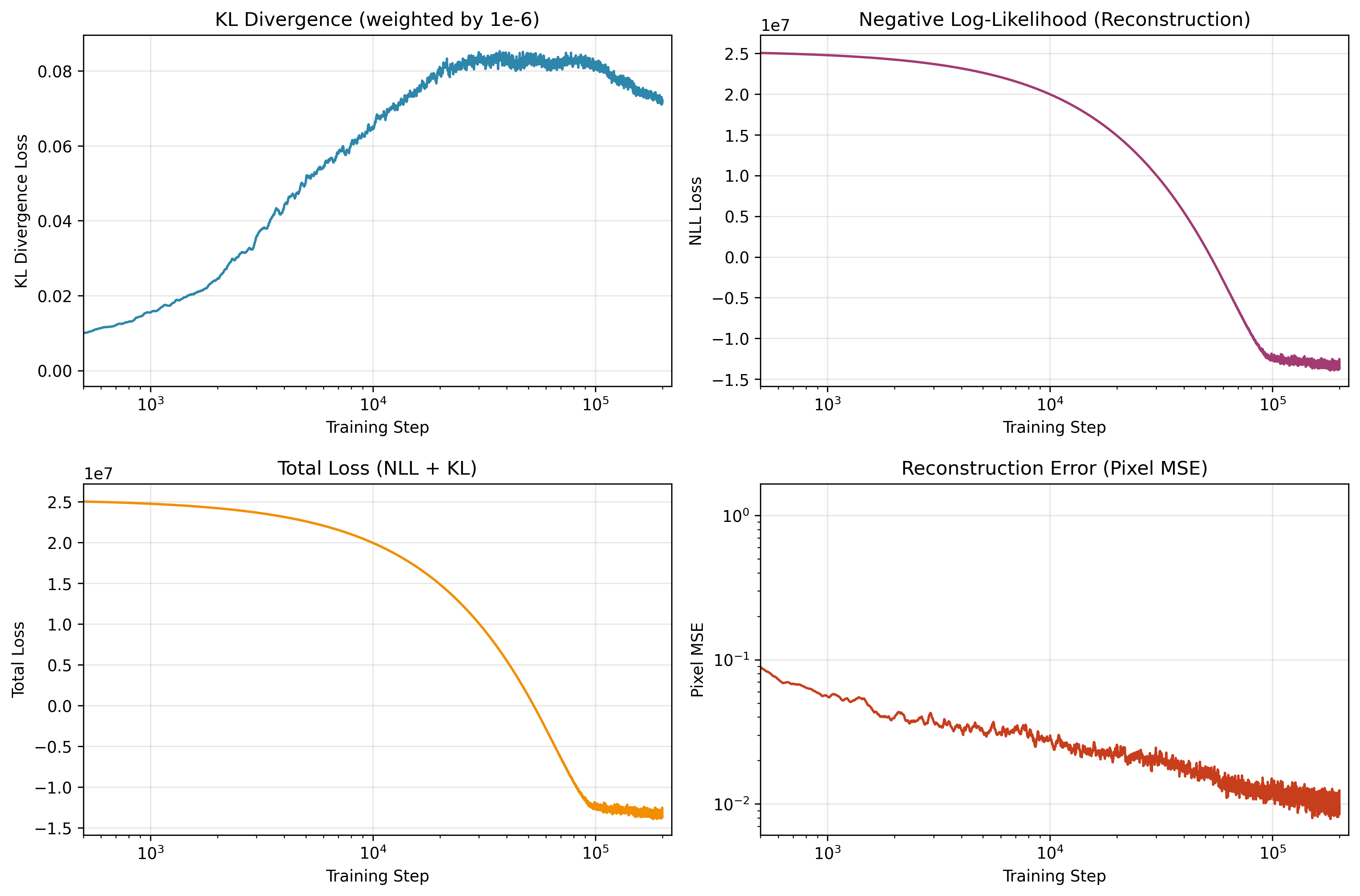}
\caption{\textbf{Base model training dynamics.} Training curves showing KL divergence loss (top left), negative log-likelihood reconstruction loss (top right), total loss (bottom left), and pixel-wise reconstruction error (bottom right) over 200,000 training steps for the unsupervised VAE baseline. All x-axes use log scale; reconstruction error (bottom right) uses log scale on y-axis. The model converges around 150,000 steps with final pixel MSE of 0.0096.}
\label{fig:base_training_dynamics}
\end{figure}

\begin{figure}[ht]
\centering
\includegraphics[width=\columnwidth]{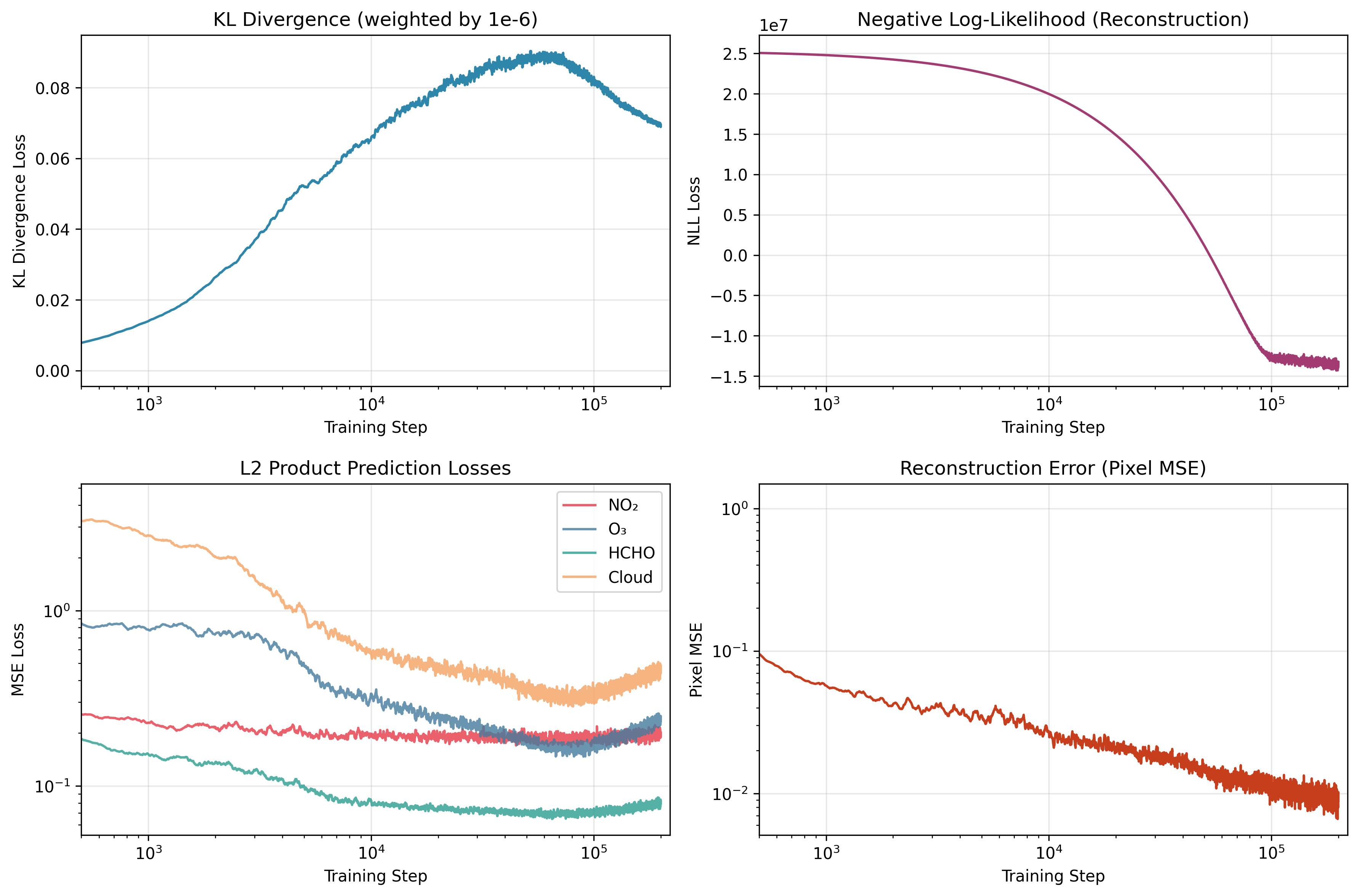}
\caption{\textbf{Latent supervised model training dynamics.} Training curves for the multi-task latent supervised VAE showing KL divergence (top left), reconstruction loss (top right), L2 product prediction losses (bottom left), and pixel reconstruction error (bottom right) over 200,000 training steps. The L2 prediction panel shows MSE losses for NO$_2$ (purple), O$_3$ (blue), HCHO (teal), and cloud fraction (orange). All axes use log scale. The supervised losses converge to final values: NO$_2$ MSE = 0.198, O$_3$ MSE = 0.234, HCHO MSE = 0.081, cloud MSE = 0.463, pixel MSE = 0.010.}
\label{fig:supervised_training_dynamics}
\end{figure}

We train using AdamW \citep{loshchilov2019decoupled} with learning rate $\alpha = 10^{-4}$, momentum $\beta_1=0.9$, $\beta_2=0.95$, weight decay $\lambda=0.05$, and gradient clipping (max norm 1.0). Training runs for 200,000 steps with batch size 32, sampling tiles randomly from a maintained buffer of 500 examples. Validation is performed every 50 steps on 100 held-out samples. Checkpoints are saved every 5,000 steps. All random seeds are set to 42, and the train/val split is 70/30 based on hashed filenames. Training converges at approximately 150,000 steps (41 hours on NVIDIA A100, peak memory 25 GB). Figure \ref{fig:base_training_dynamics} shows the training dynamics for the base unsupervised VAE model, while Figure \ref{fig:supervised_training_dynamics} displays the training curves for the latent supervised variant with additional L2 prediction losses.

\begin{table}[h]
\centering
\caption{Complete training hyperparameters.}
\small
\begin{tabular}{ll}
\hline
\textbf{Hyperparameter} & \textbf{Value} \\
\hline
\multicolumn{2}{l}{\textit{Optimizer Settings}} \\
Optimizer & AdamW \\
Learning Rate ($\alpha$) & $1 \times 10^{-4}$ \\
Momentum ($\beta_1, \beta_2$) & [0.9, 0.95] \\
Weight Decay ($\lambda$) & 0.05 \\
Gradient Clip Norm & 1.0 \\
\hline
\multicolumn{2}{l}{\textit{Loss Configuration}} \\
Reconstruction Loss & L1 (MAE) \\
KL Divergence Weight & $1 \times 10^{-6}$ \\
Learned Log-Variance (init) & 6.0 \\
\hline
\multicolumn{2}{l}{\textit{Training Schedule}} \\
Total Steps & 200,000 \\
Batch Size & 32 \\
Validation Frequency & 50 steps \\
Checkpoint Frequency & 5,000 steps \\
\hline
\multicolumn{2}{l}{\textit{Data Configuration}} \\
Train/Val Split & 70/30 \\
Training Buffer Size & 500 samples \\
Validation Buffer Size & 100 samples \\
Random Seed & 42 \\
\hline
\multicolumn{2}{l}{\textit{Computational Resources}} \\
Training Time & 41 hours \\
GPU & NVIDIA A100 (40GB) \\
Peak GPU Memory & $\sim$25 GB \\
\hline
\end{tabular}
\label{tab:hyperparameters}
\end{table}

\section{Probing Methodology}
\label{app:probing}

After training the VAE to compress TEMPO hyperspectral radiance data, we evaluate whether atmospheric composition information is preserved in the learned latent representations through a two-stage probing framework. We freeze the VAE encoder weights and train supervised regression models (probes) to predict Level-2 atmospheric products from the latent space. This approach tests whether the unsupervised compression has implicitly captured atmospheric signals relevant for trace gas and cloud property retrievals.

The probing experiments address a fundamental question: does dimensionality reduction through unsupervised learning preserve physically meaningful atmospheric information? By comparing linear and nonlinear probe performance, we assess whether atmospheric products are encoded linearly in the latent space or require more complex extraction mechanisms. The dataset for probing consists of spatially aligned pairs of VAE latent representations (extracted from Level-1 radiance) and normalized Level-2 product values, with samples drawn uniformly from valid (non-NaN) pixels across all training files. We use an 80/20 train/test split with fixed random seed (42) for reproducibility. All probes are trained to minimize mean squared error (MSE) between predicted and ground-truth normalized product values.

\subsection{Linear Probing}

Linear probes provide a baseline assessment of whether atmospheric information is directly accessible through linear combinations of latent channels. Each probe consists of a single linear layer mapping from the 32 latent channels to a scalar output representing one atmospheric product: $\hat{y} = W^\top z + b$, where $z \in \mathbb{R}^{32}$ is the latent vector at a spatial location, $W \in \mathbb{R}^{32}$ are learned weights, $b \in \mathbb{R}$ is a bias term, and $\hat{y}$ is the predicted normalized product value.

Linear probes are trained for 100 epochs using the AdamW optimizer with learning rate 0.001 and weight decay 0.01 (L2 regularization). The batch size is 512, and we sample 2000 valid pixels per file for training. The simplicity of linear probes makes them computationally efficient ($<$1 minute training time per product) and provides interpretability—the learned weight vector $W$ reveals which latent channels contribute most strongly to each atmospheric product prediction. Poor linear probe performance suggests that relevant information exists in the latent space but requires nonlinear transformations for extraction, motivating the use of MLP probes.

\subsection{MLP Probing}

To test whether atmospheric signals are encoded nonlinearly in the VAE latent space, we employ multi-layer perceptron (MLP) probes capable of learning complex nonlinear mappings. The MLP architecture consists of three fully-connected layers with dimensions [32 $\rightarrow$ 512 $\rightarrow$ 512 $\rightarrow$ 1], using ReLU activations between layers and dropout (probability 0.1) for regularization. The forward pass computes:
\begin{align*}
h_1 &= \text{ReLU}(W_1 z + b_1) \\
h_2 &= \text{Dropout}(\text{ReLU}(W_2 h_1 + b_2)) \\
\hat{y} &= W_3 h_2 + b_3
\end{align*}
where $h_1, h_2 \in \mathbb{R}^{512}$ are hidden representations and all other notation follows the linear case.

MLP probes train for up to 2000 epochs with early stopping (patience=10 epochs on validation loss) using AdamW optimizer (learning rate 0.001, weight decay 0.01). We sample 1000 valid pixels per file with batch size 512. The substantially deeper architecture (approximately 280K parameters per probe compared to 33 for linear probes) provides capacity to learn arbitrary nonlinear functions of the latent representation, at the cost of increased training time ($\sim$3-5 minutes per product) and reduced interpretability. Significant performance gains from MLP over linear probes indicate that atmospheric information is present but encoded nonlinearly, suggesting that either (1) the VAE's unsupervised objective does not naturally align latent dimensions with atmospheric products, or (2) the products themselves depend nonlinearly on the observed radiances. Conversely, minimal improvement suggests fundamental limitations in how much information about that product is preserved during compression.

\begin{table}[h]
\centering
\caption{Training configurations for linear and MLP probes.}
\begin{tabular}{lll}
\hline
\textbf{Parameter} & \textbf{Linear Probe} & \textbf{MLP Probe} \\
\hline
Architecture & [32 → 1] & [32 → 512 → 512 → 1] \\
Activation & None & ReLU \\
Dropout & 0.0 & 0.1 \\
Optimizer & AdamW & AdamW \\
Learning Rate & 0.001 & 0.001 \\
Weight Decay & 0.01 & 0.01 \\
Batch Size & 512 & 512 \\
Max Epochs & 100 & 2000 \\
Early Stopping & Yes (patience=10) & Yes (patience=10) \\
Pixels per File & 2000 & 1000 \\
Total Samples & 140,000 & 70,000 \\
Train/Test Split & 80/20 & 80/20 \\
\hline
\end{tabular}
\label{tab:probe_config}
\end{table}

\section{Additional Results}
\label{app:additional_results}

\subsection{Probe Training Dynamics}
\label{app:probe_training}

\begin{figure*}[h!]
\centering
\includegraphics[width=0.46\textwidth]{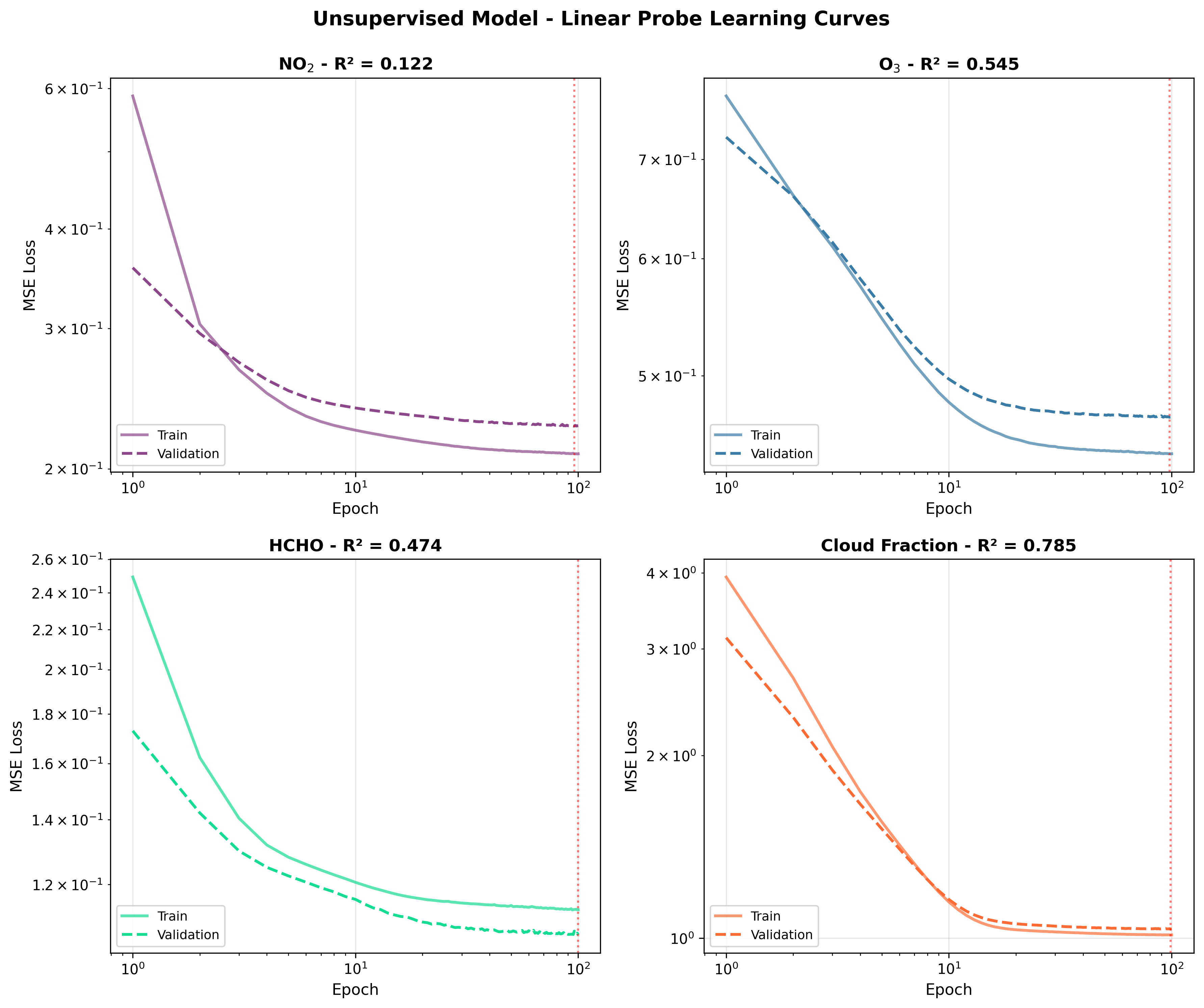}
\hspace{1cm}
\includegraphics[width=0.46\textwidth]{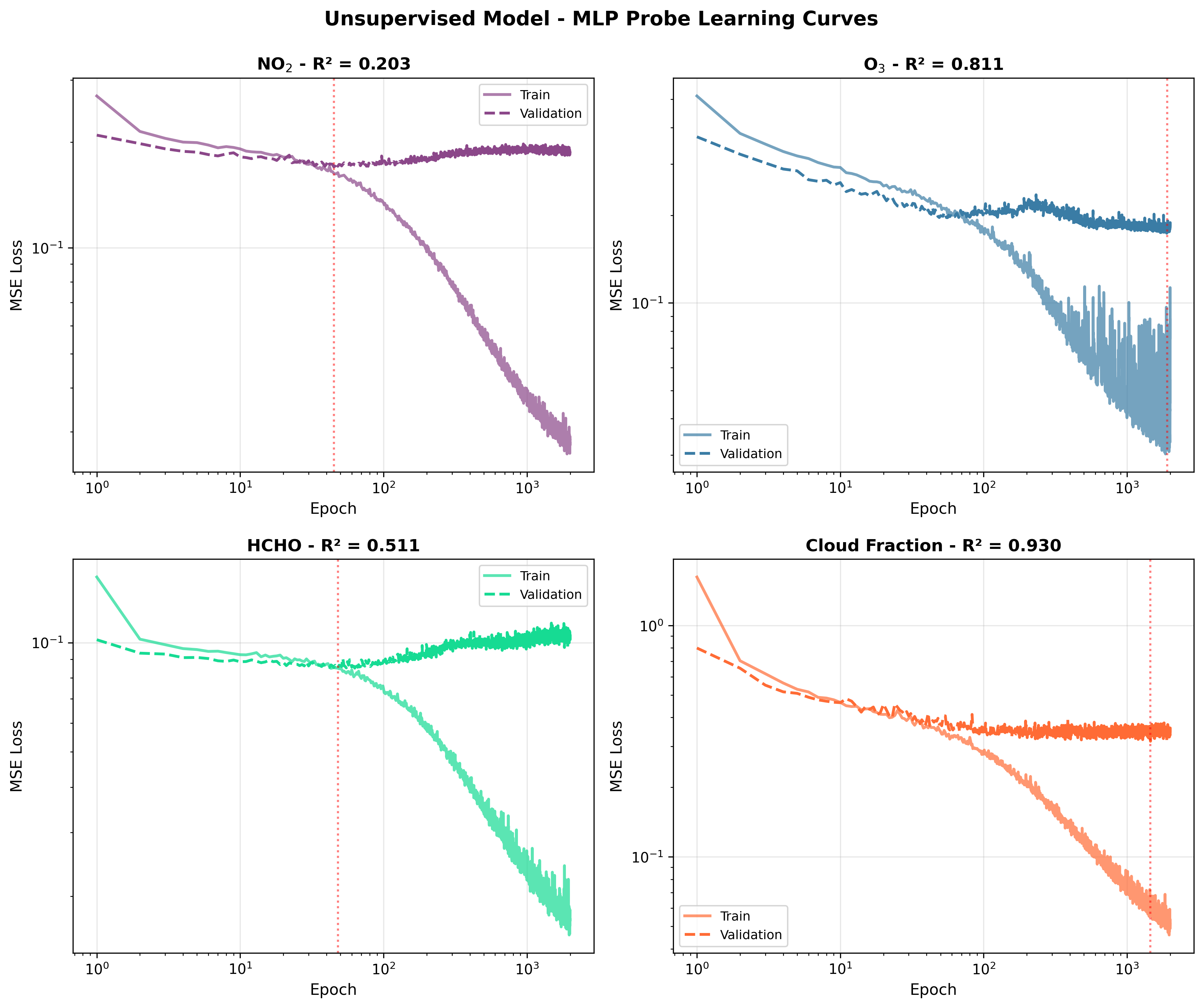}
\caption{\textbf{Unsupervised VAE probe training dynamics.} Learning curves for (left) linear probes trained for 100 epochs and (right) MLP probes ([32→512→512→1] architecture) trained for up to 2000 epochs on the base unsupervised VAE latent representations. Each panel shows training (solid) and validation (dashed) MSE loss for all four atmospheric products: NO$_2$ (purple), O$_3$ (blue), HCHO (teal), and cloud fraction (orange). Red dotted vertical lines indicate the epoch with minimum validation loss. MLP probes substantially outperform linear probes, particularly for cloud fraction (R$^2$: 0.785→0.930) and total ozone (R$^2$: 0.545→0.811), demonstrating that atmospheric information is encoded nonlinearly in the compressed representation.}
\label{fig:unsupervised_probe_learning}
\end{figure*}

\begin{figure*}[h!]
\centering
\includegraphics[width=0.46\textwidth]{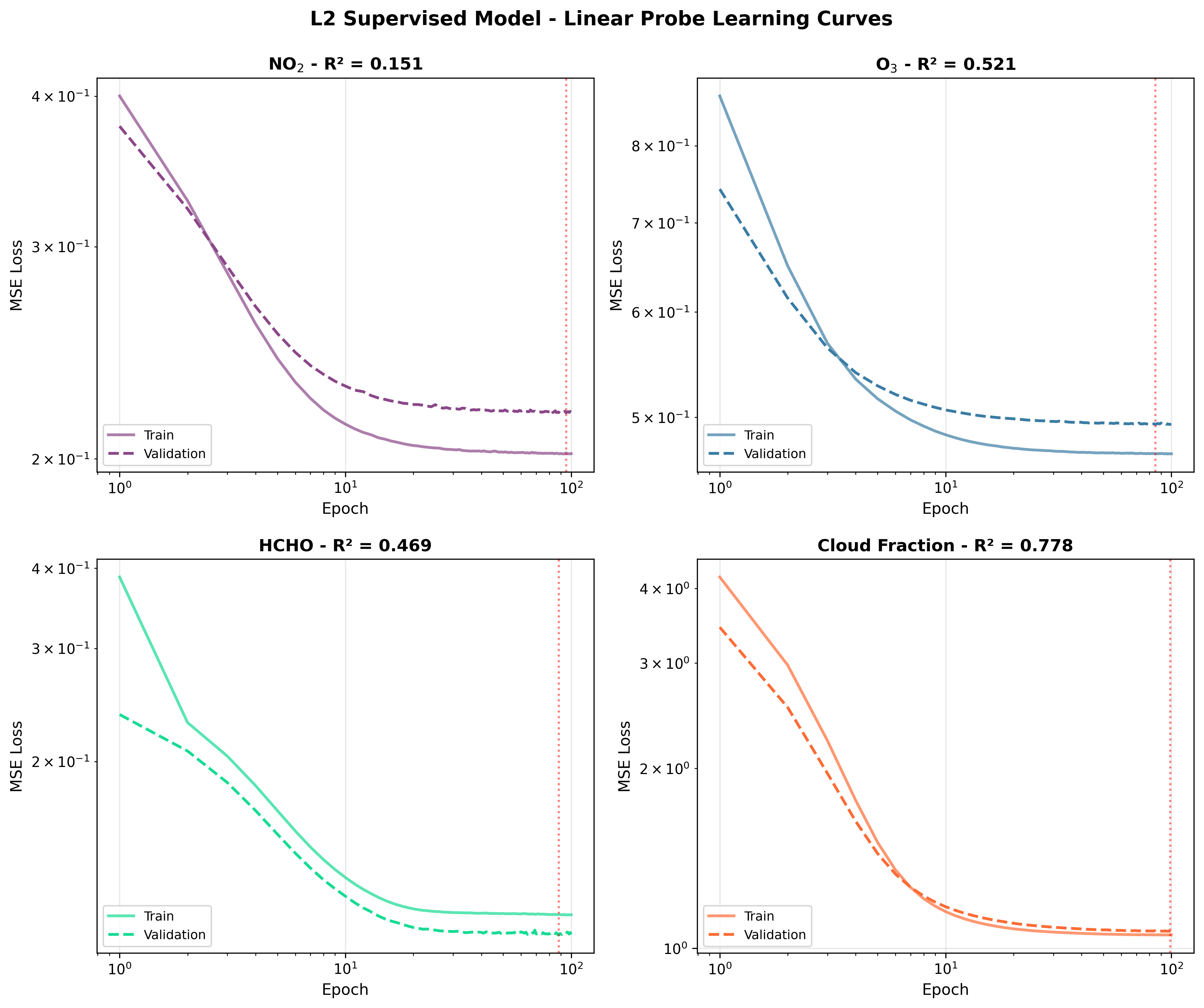}
\hspace{1cm}
\includegraphics[width=0.46\textwidth]{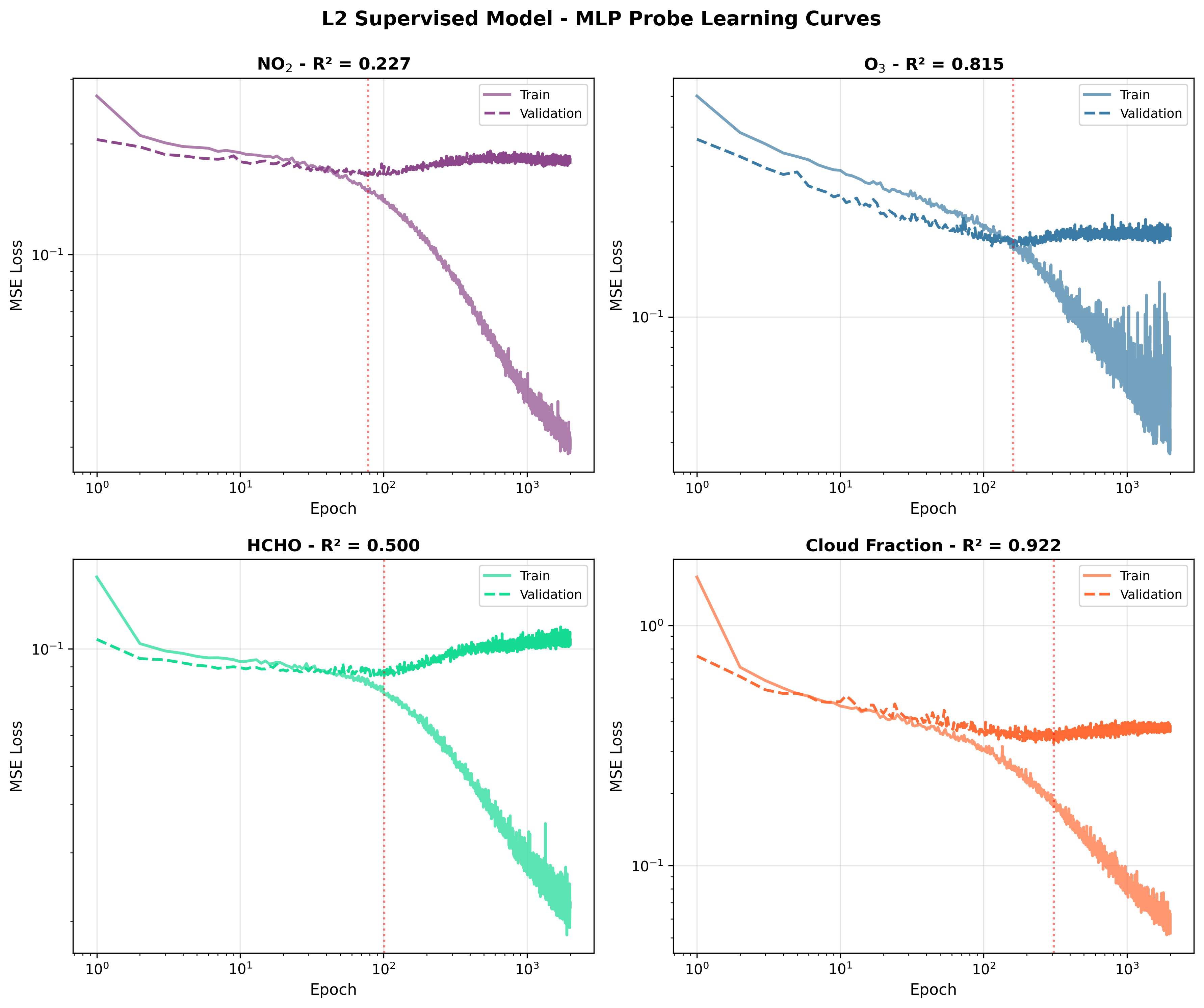}
\caption{\textbf{L2-supervised VAE probe training dynamics.} Learning curves for (left) linear probes and (right) MLP probes trained on the L2-supervised VAE latent representations. Despite explicit supervision during VAE training to predict L2 products, the training dynamics and final performance are nearly identical to the unsupervised model (Figure \ref{fig:unsupervised_probe_learning}). Linear probes achieve R$^2$ scores of 0.778 (cloud), 0.521 (O$_3$), 0.469 (HCHO), and 0.151 (NO$_2$), while MLP probes reach 0.922 (cloud), 0.815 (O$_3$), 0.500 (HCHO), and 0.227 (NO$_2$). The similarity indicates that the reconstruction objective already captures atmospheric structure effectively.}
\label{fig:supervised_probe_learning}
\end{figure*}

To assess the convergence and optimization behavior of our probing framework, we provide complete training dynamics for both linear and MLP probes applied to the unsupervised and L2-supervised VAE models. Figures \ref{fig:unsupervised_probe_learning} and \ref{fig:supervised_probe_learning} show the MSE loss curves during probe training for all four atmospheric products.

The unsupervised base VAE model (Figure \ref{fig:unsupervised_probe_learning}) shows distinct training dynamics between linear and MLP probes. Linear probes converge rapidly within 100 epochs with smooth monotonic decrease, achieving moderate performance (cloud fraction R$^2 \approx 0.78$, NO$_2$ R$^2 \approx 0.12$). MLP probes require substantially more training iterations (up to 2000 epochs) but achieve significantly better performance, particularly for cloud fraction (R$^2 \approx 0.93$) and total ozone (R$^2 \approx 0.81$). The training curves reveal product-specific patterns: NO$_2$ and HCHO exhibit initial plateaus before rapid descent around epoch 100, suggesting these products require the MLP to learn complex nonlinear transformations. In contrast, O$_3$ and cloud fraction show steady improvement from the start, indicating that their nonlinear encoding is more readily accessible.

The L2-supervised VAE model (Figure \ref{fig:supervised_probe_learning}) exhibits remarkably similar training dynamics and final performance compared to the unsupervised model. Both linear and MLP probes achieve nearly identical R$^2$ scores across all products, with cloud fraction reaching R$^2 \approx 0.92$ and total ozone R$^2 \approx 0.81$ using MLP probes. This similarity demonstrates that explicit L2 supervision during VAE training provides minimal benefit for post-hoc atmospheric product extraction—the reconstruction objective alone captures sufficient atmospheric structure in the latent representations.

\subsection{Additional Reconstruction Examples}

\begin{figure*}[h!]
\centering
\includegraphics[width=0.6\textwidth]{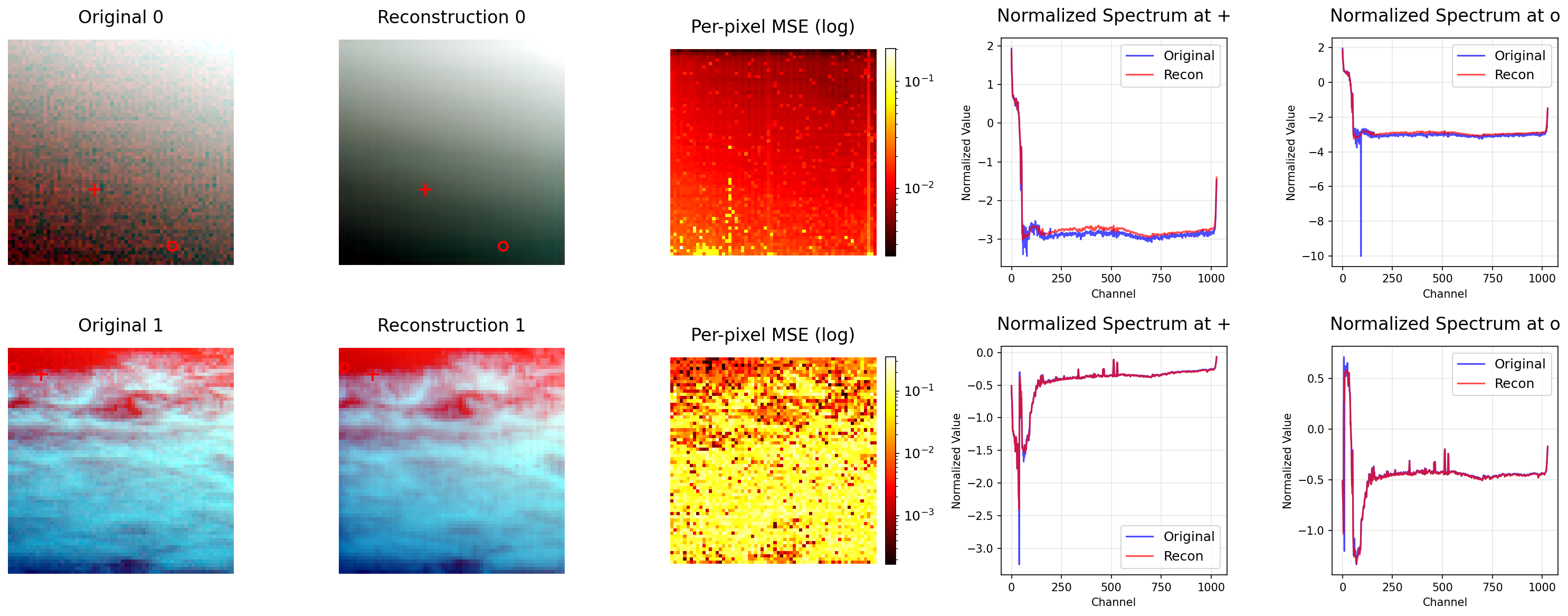}\\[0.5em]
\includegraphics[width=0.6\textwidth]{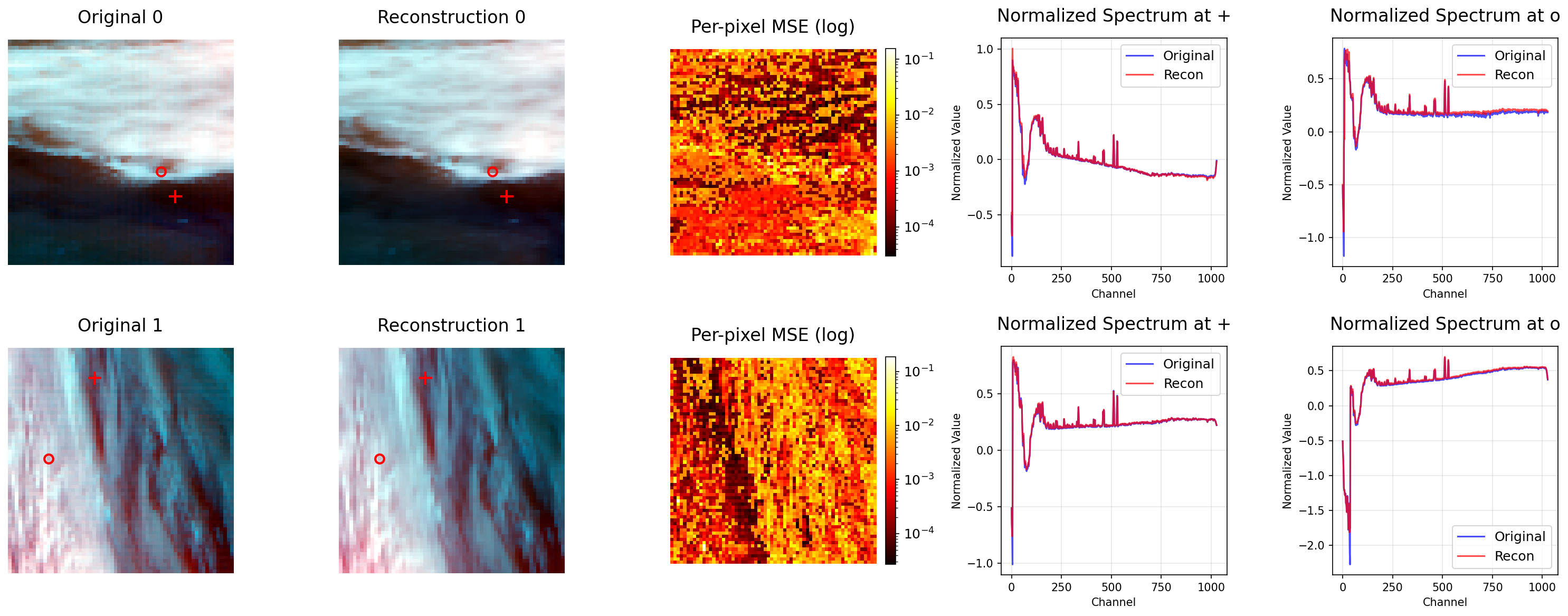}\\[0.5em]
\includegraphics[width=0.6\textwidth]{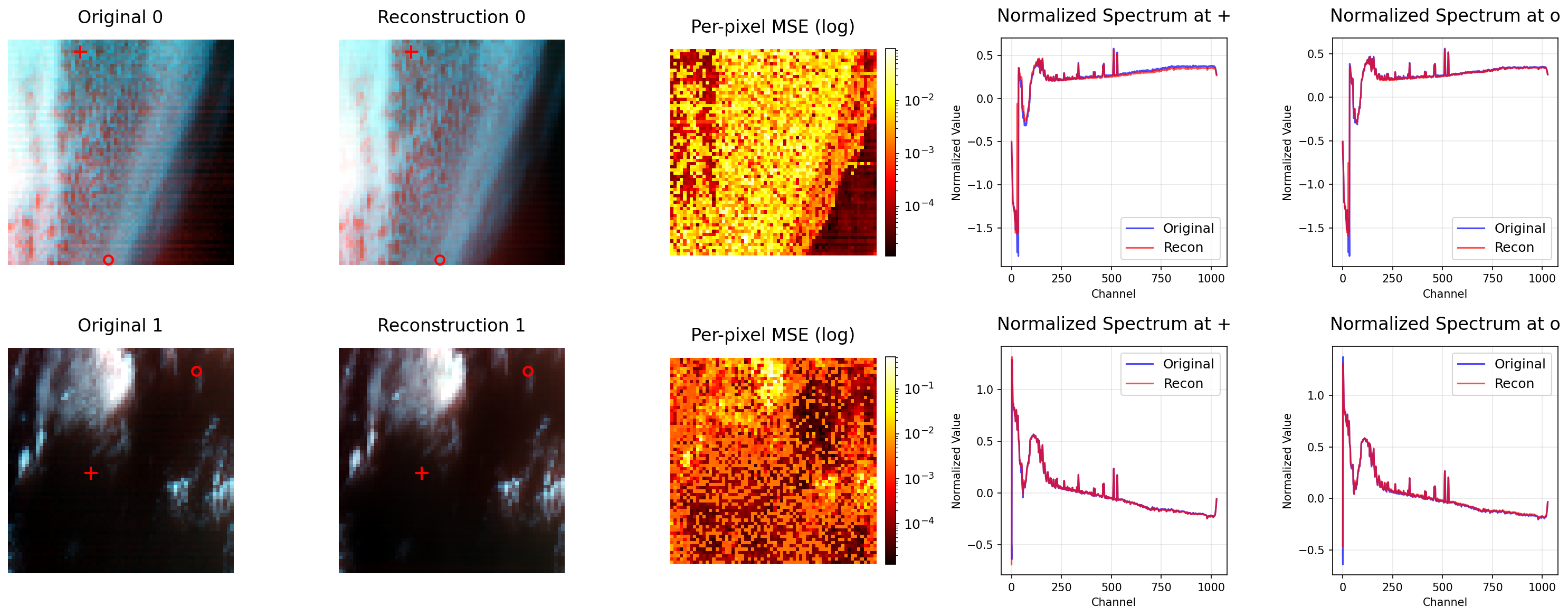}\\[0.5em]
\includegraphics[width=0.6\textwidth]{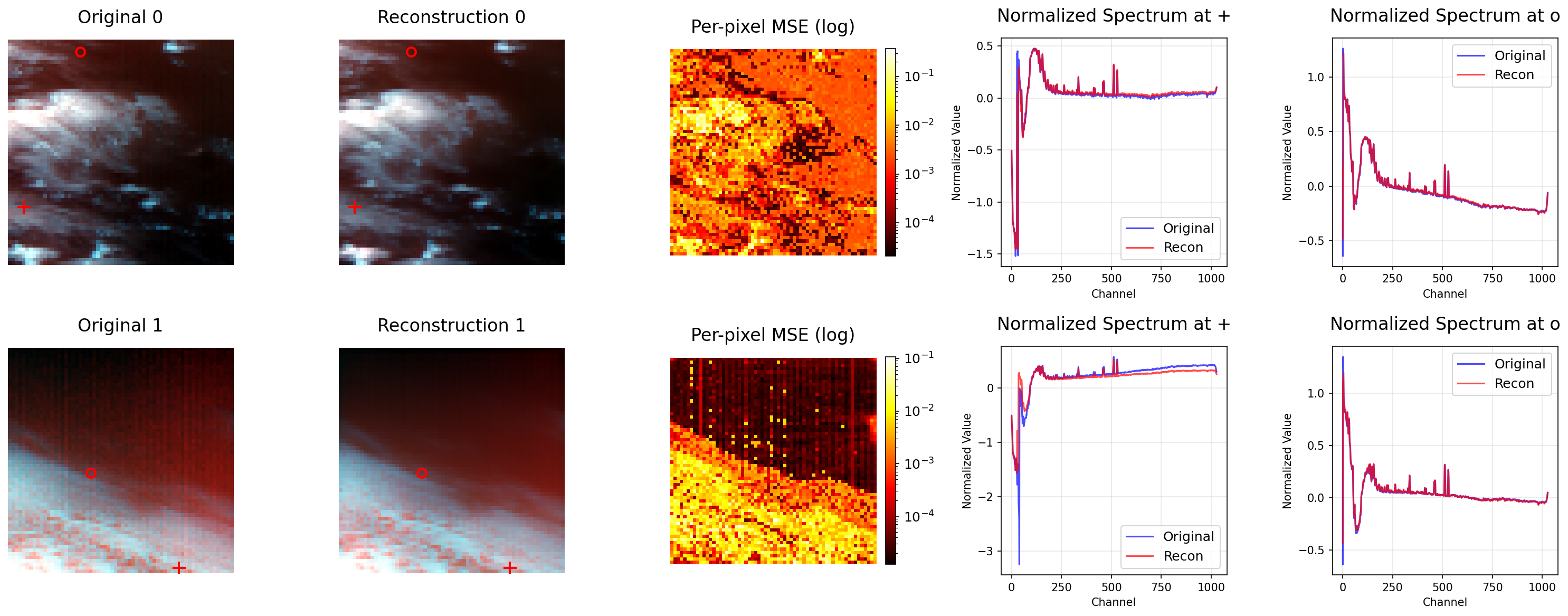}\\[0.5em]
\includegraphics[width=0.6\textwidth]{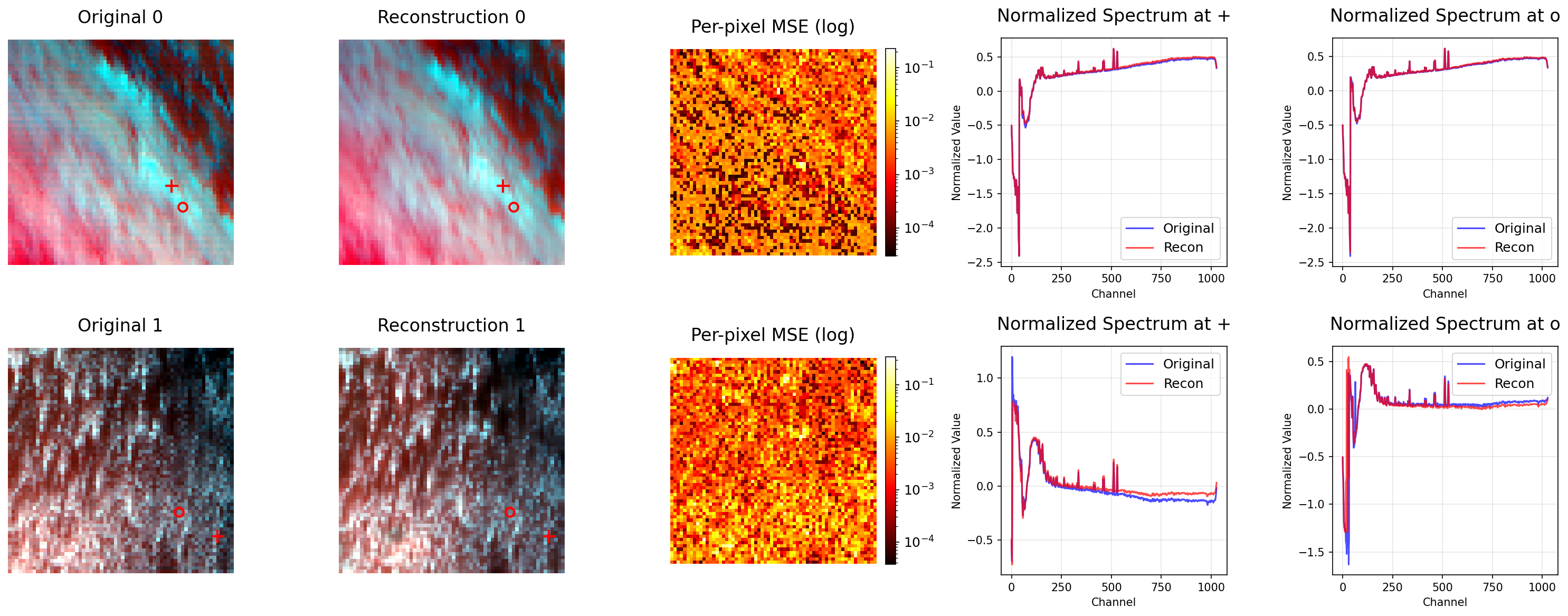}
\caption{\textbf{Base VAE reconstructions.} Five randomly selected validation samples after 200,000 training steps. Each row: (1) original L1 radiance, (2) reconstruction, (3) per-pixel MSE (log scale), (4-5) spectra at two locations.}
\label{fig:reconstruction_examples}
\end{figure*}

\begin{figure*}[h!]
\centering
\includegraphics[width=0.6\textwidth]{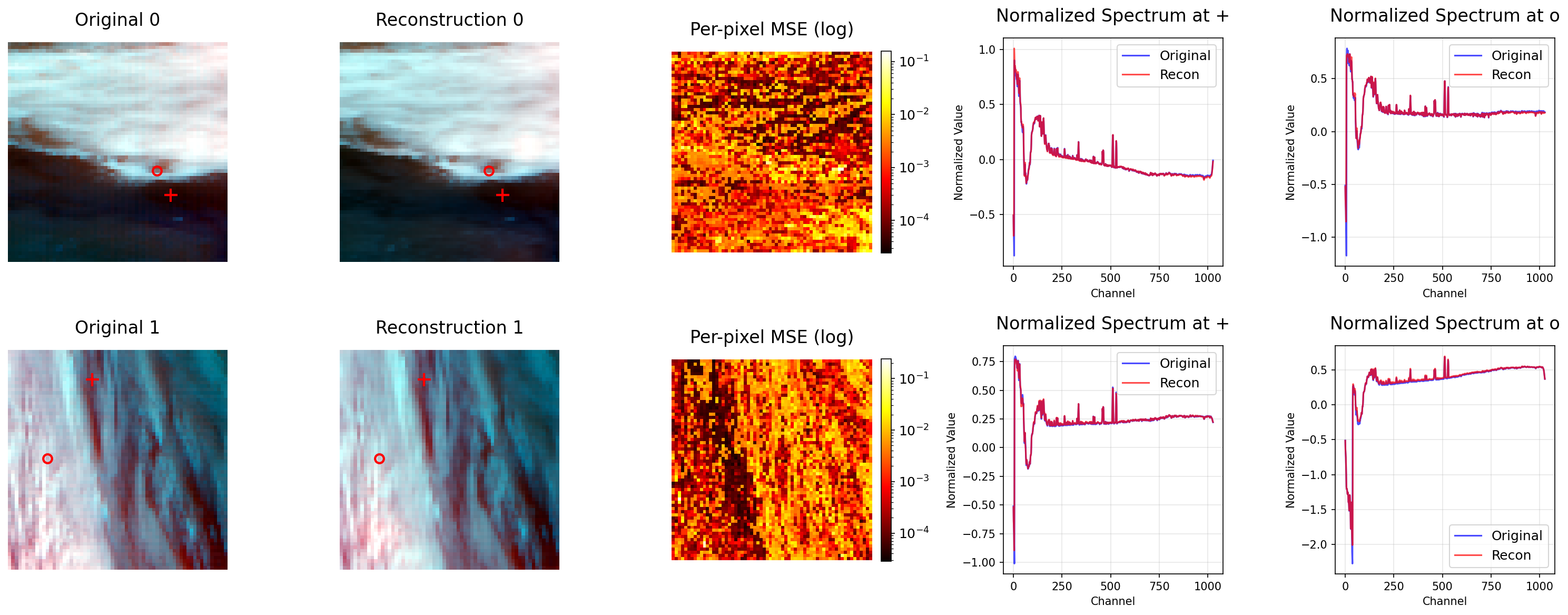}\\[0.5em]
\includegraphics[width=0.6\textwidth]{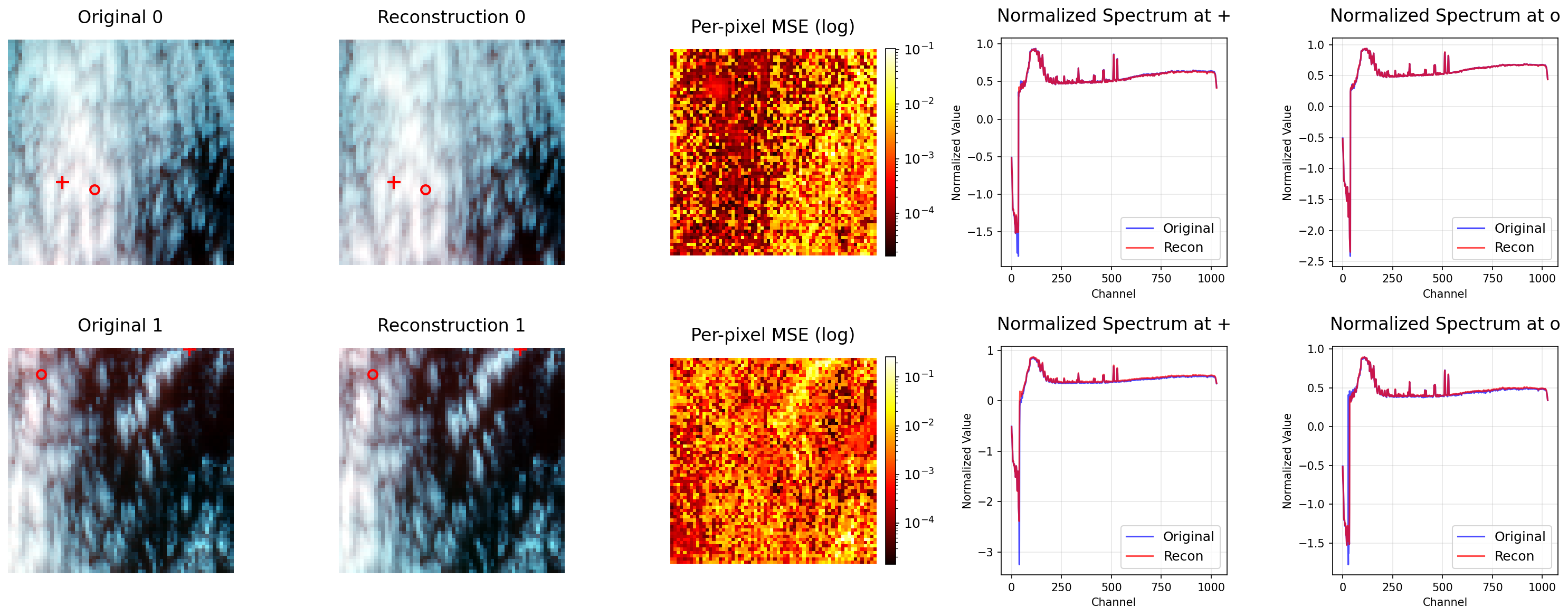}\\[0.5em]
\includegraphics[width=0.6\textwidth]{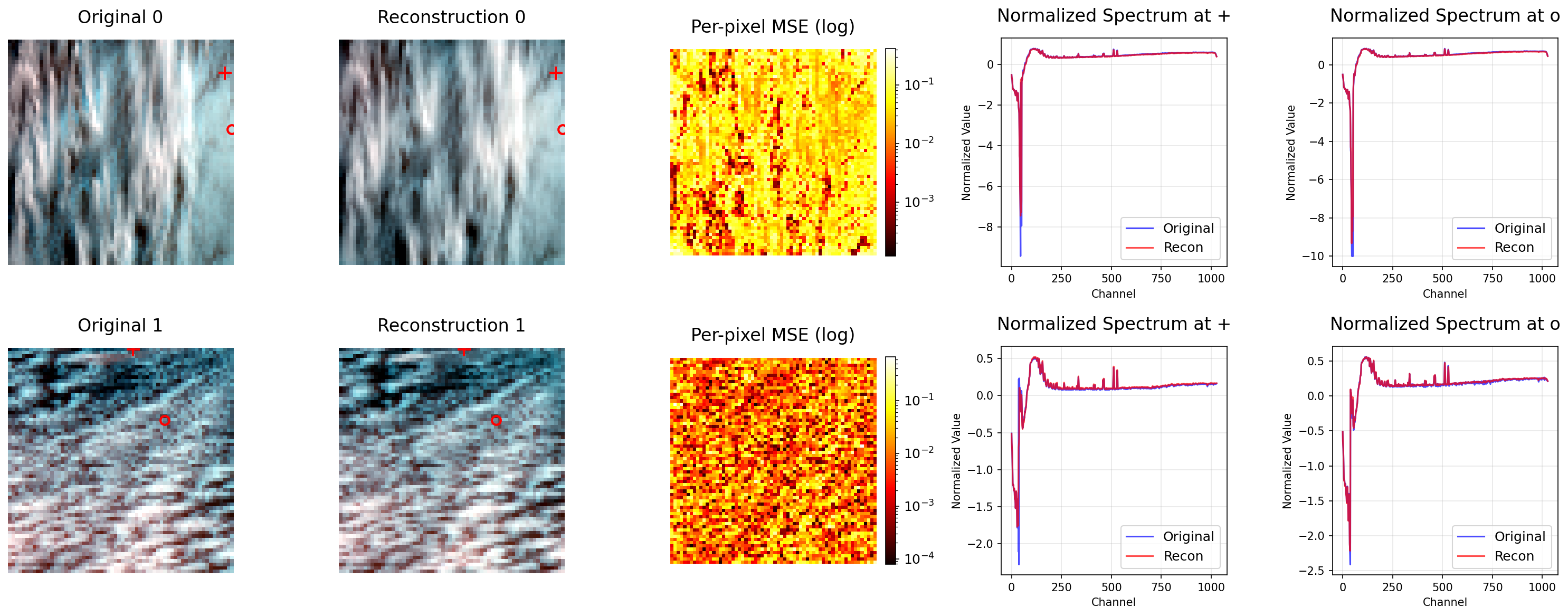}\\[0.5em]
\includegraphics[width=0.6\textwidth]{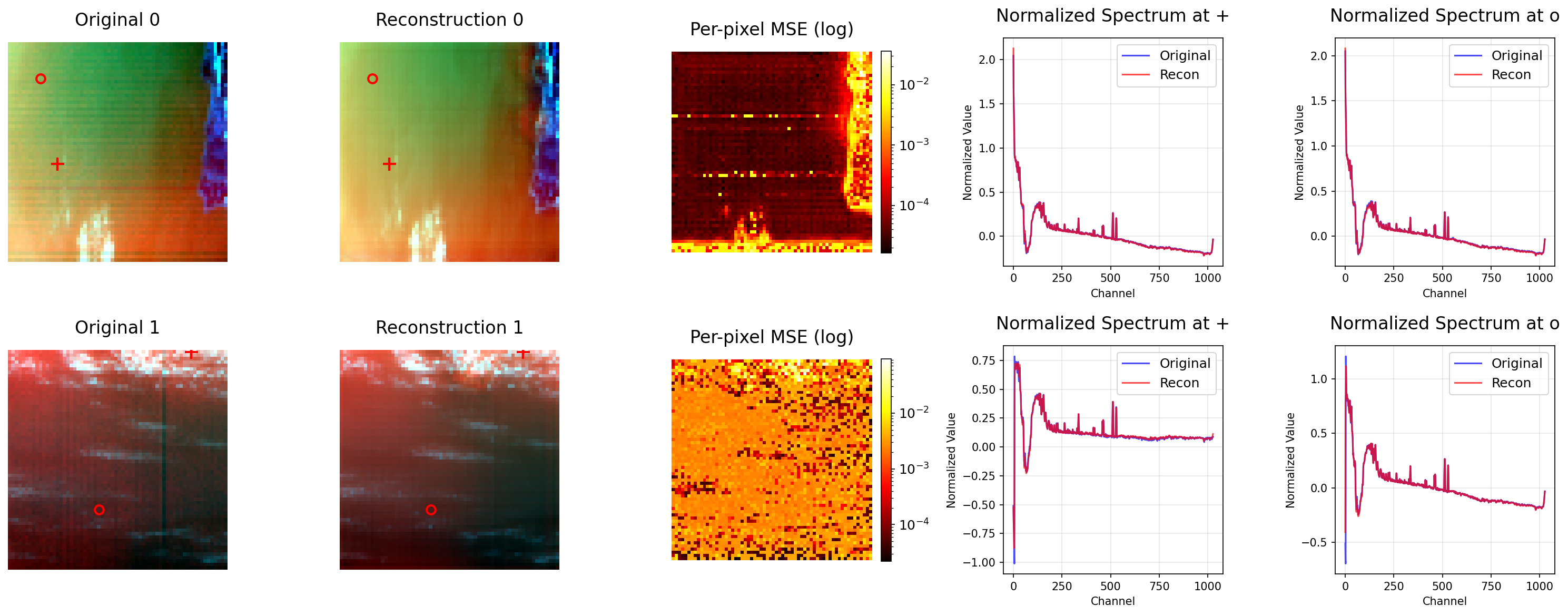}\\[0.5em]
\includegraphics[width=0.6\textwidth]{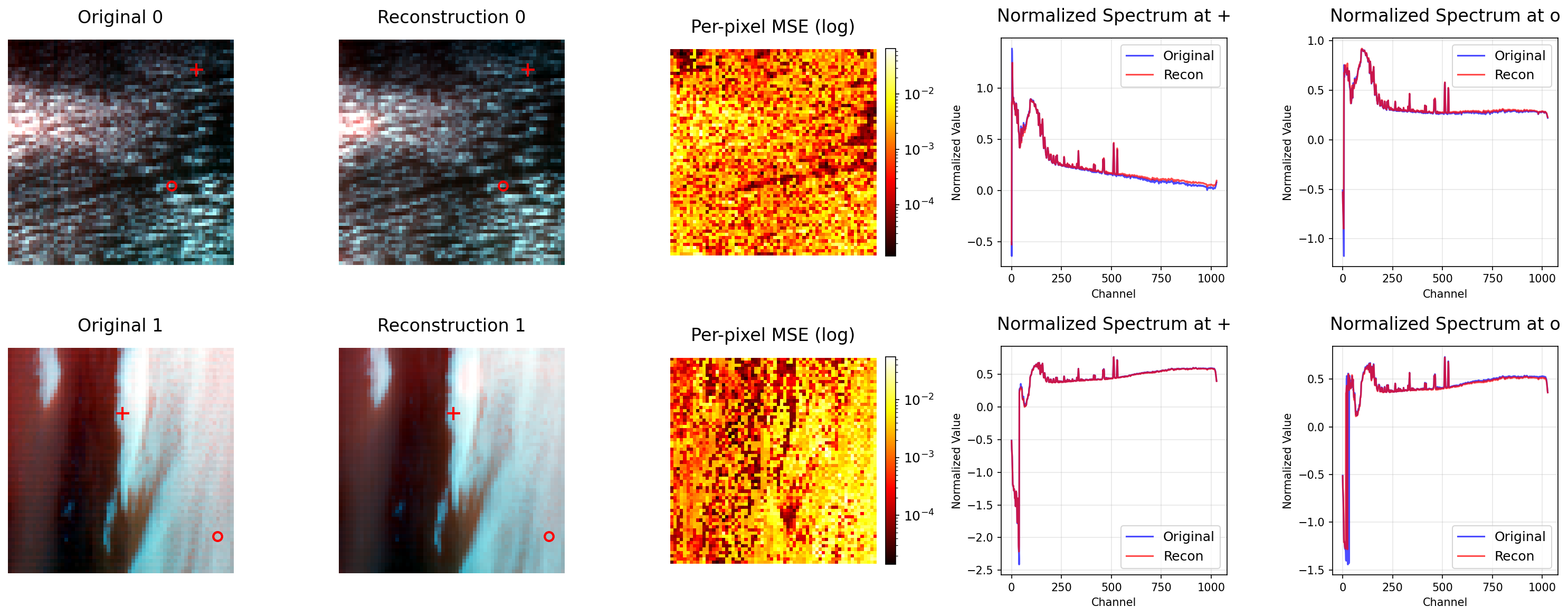}
\caption{\textbf{Latent supervised VAE reconstructions.} Five randomly selected validation samples after 200,000 training steps. The first panel corresponds to the same validation sample shown in Figure 2 (main text). Same format as Figure \ref{fig:reconstruction_examples}.}
\label{fig:reconstruction_examples_supervised}
\end{figure*}

Figure \ref{fig:reconstruction_examples} shows five randomly selected validation samples from the base VAE model, while Figure \ref{fig:reconstruction_examples_supervised} shows five randomly selected validation samples from the latent supervised VAE model. The first panel in Figure \ref{fig:reconstruction_examples_supervised} corresponds to the same validation sample shown in the main Figure 2.

\section{Code and Data Availability}
\label{app:availability}

All code and configuration files will be available after the peer review process.

Raw TEMPO satellite data is publicly available through NASA's Earthdata portal: \texttt{https://www.earthdata.nasa.gov/}
or \texttt{https://asdc.larc.nasa.gov/project/TEMPO}.

\section{Computational Requirements}
\label{app:compute}

All experiments were conducted on NVIDIA A100-SXM4-40GB GPUs using PyTorch \citep{paszke2019pytorch}. VAE training required 41 hours for 200,000 steps with batch size 32, using peak GPU memory of $\sim$25GB for the 27.3M parameter model processing input tensors of shape $[32 \times 1028 \times 64 \times 64]$. Probe training took $<$5 minutes per component. Total storage requirement is approximately 220GB including 102GB raw TEMPO L1B files (70 files, Los Angeles region, January 2025), 51GB processed tiles, 51GB tiles with L2 products, 15GB individual L2 products, and model checkpoints ($\sim$1GB per checkpoint).

\end{document}